\documentclass[11pt]{article}
\usepackage[labelfont=bf]{caption}
\usepackage{url,hyperref,lineno,microtype,subcaption}
\usepackage[onehalfspacing]{setspace}
\usepackage{amsmath, amsfonts}
\usepackage{graphicx}
\usepackage{dcolumn}
\usepackage{bm}
\usepackage{tabularx}
\usepackage{color}
\usepackage[english]{babel}

\newtheorem{thm}{Theorem}
\newtheorem{lem}{Lemma}

\newtheorem{defn}{Definition}

\begin{document}

\centerline{\LARGE \bf Exact heat kernel on a hypersphere and}
\centerline{\LARGE\bf its applications in kernel SVM} 

\bigskip

\centerline{Chenchao Zhao\,$^{1,2}$, and Jun S.~Song\,$^{1,2,*}$}
\bigskip
{\noindent $^{1}$Department of Physics,
University of Illinois at Urbana-Champaign,
Urbana, IL 61801, USA \\
$^{2}$Carl R.~Woese Institute for Genomic Biology,
University of Illinois at Urbana-Champaign,
Urbana, IL 61801, USA }

\medskip

{\noindent $^{*}$ Correspondence: songj@illinois.edu}

\begin{abstract}

  Many contemporary statistical learning methods assume a Euclidean
  feature space.  This paper presents a method for defining similarity
  based on hyperspherical geometry and shows that it often improves
  the performance of support vector machine compared to other
  competing similarity measures.  Specifically, the idea of using heat
  diffusion on a hypersphere to measure similarity has been previously
  proposed and tested by \cite{Lafferty:2005uy}, demonstrating
  promising results based on a heuristic heat kernel obtained from the
  zeroth order parametrix expansion; however, how well this heuristic
  kernel agrees with the exact hyperspherical heat kernel remains
  unknown.  This paper presents a higher order parametrix expansion of
  the heat kernel on a unit hypersphere and discusses several problems
  associated with this expansion method.  We then compare the
  heuristic kernel with an exact form of the heat kernel expressed in
  terms of a uniformly and absolutely convergent series in
  high-dimensional angular momentum eigenmodes.  Being a natural
  measure of similarity between sample points dwelling on a
  hypersphere, the exact kernel often shows superior performance in
  kernel SVM classifications applied to text mining, tumor somatic
  mutation imputation, and stock market analysis.

\end{abstract}

\section{Introduction}

As the techniques for analyzing large data sets continue to grow,
diverse quantitative sciences -- including computational biology,
observation astronomy, and high energy physics -- are becoming
increasingly data driven. Moreover, modern business decision making
critically depends on quantitative analyses such as community
detection and consumer behavior prediction. Consequently, statistical
learning has become an indispensable tool for modern data
analysis. Data acquired from various experiments are usually organized
into an $n\times m$ matrix, where the number $n$ of features typically
far exceeds the number $m$ of samples. In this view, the $m$ samples,
corresponding to the columns of the data matrix, are naturally
interpreted as points in a high-dimensional feature space
$\mathbb{R}^{n}$.  Traditional statistical modeling approaches often
lose their power when the feature dimension is high. 
To ameliorate this problem, Lafferty and Lebanon
proposed a multinomial interpretation of non-negative feature vectors
and an accompanying transformation of the multinomial simplex to a
hypersphere, demonstrating that using the heat kernel on this
hypersphere may improve the performance of kernel support vector
machine (SVM)
\cite{Lafferty:2005uy,Hastie:2013fd,Evgeniou:2001hc,Boser:1992fz,Cortes:1995fs,
  Freund:1999dh,Guyon:1993ub}. Despite the interest that this idea has
attracted, only approximate heat kernel is known to date. We here
present an exact form of the heat kernel on a hypersphere of arbitrary
dimension and study its performance in kernel SVM classifications of
text mining, genomic, and stock price data sets.

To date, sparse data clouds have been extensively analyzed in the flat
Euclidean space endowed with the $L^{2}$-norm using traditional
statistical learning algorithms, including KMeans, hierarchical
clustering, SVM, and neural network \cite{Hastie:2013fd,
  Kaufman:2009dk,
  Evgeniou:2001hc,Boser:1992fz,Cortes:1995fs,Freund:1999dh,Guyon:1993ub};
however, the flat geometry of the Euclidean space often poses severe
challenges in clustering and classification problems when the data
clouds take non-trivial geometric shapes or class labels are spatially
mixed. Manifold learning and kernel-based embedding methods attempt to
address these challenges by estimating the intrinsic geometry of a
putative submanifold from which the data points were sampled and by
embedding the data into an abstract Hilbert space using a nonlinear
map implicitly induced by the chosen kernel, respectively
\cite{Anonymous:U64WGXbS,Aronszajn:1950bl,Paulsen:2016wz}.  The
geometry of these curved spaces may then provide novel information
about the structure and organization of original data points.

Heat equation on the data submanifold or transformed feature space
 offers an especially attractive idea of measuring similarity between
data points by using the physical model of diffusion of relatedness
(``heat'') on curved space, where the diffusion process is driven
by the intrinsic geometry of the underlying space. Even though such
diffusion process has been successfully approximated as a discrete-time,
discrete-space random walk on complex networks, its continuous formulation
is rarely analytically solvable and usually requires complicated asymptotic
expansion techniques from differential geometry \cite{Berger:wXGLCovN}.
An analytic solution, if available, would thus provide a valuable
opportunity for comparing its performance with approximate asymptotic
solutions and rigorously testing the power of heat diffusion for geometric
data analysis.

Given that a Riemannian manifold of dimension $d$ is locally
homeomorphic to $\mathbb{R}^{d}$, and that the heat kernel is a
solution to the heat equation with a point source initial condition,
one may assume in the short diffusion time limit ($t\downarrow 0$) that most
of the heat is localized within the vicinity of the initial point and
that the heat kernel on a Riemannian manifold locally resembles the
Euclidean heat kernel. This idea forms the motivation behind the
parametrix expansion, where the heat kernel in curved space is
approximated as a product of the Euclidean heat kernel in normal
coordinates and an asymptotic series involving the diffusion time and
normal coordinates.  In particular, for a unit hypersphere, the
parametrix expansion in the limit $t\downarrow 0$ involves a modified
Euclidean heat kernel with the Euclidean distance $\left\Vert
  \mathbf{x}\right\Vert $ replaced by the geodesic arc length
$\theta$.  Computing this parametrix expansion is, however,
technically challenging; even when the computation is tractable,
applying the approximation directly to high-dimensional clustering and
classification problems may have limitations. For example, in order to
be able to group samples robustly, one needs the diffusion time $t$ to
be not too small; otherwise, the sample relatedness may be highly
localized and decay too fast away from each sample. Moreover, the
leading order term in the asymptotic series is an increasing function
of $\theta$ and diverges as $\theta$ approaches $\pi$, yielding an
incorrect conclusion that two antipodal points are highly similar. For
these reasons, the machine learning community has been using only the
Euclidean diffusion term without the asymptotic series correction; how
this resulting kernel, called the parametrix kernel
\cite{Lafferty:2005uy}, compares with the exact heat kernel on a
hypersphere remains an outstanding question, which is addressed in
this paper.

Analytically solving the diffusion equation on a Riemannian manifold
is challenging \cite{Hsu:2002vq,Varopoulos:1987vx,Berger:wXGLCovN}.
Unlike the discrete analogues \textendash{} such as spectral clustering
\cite{Ng:2002tj} and diffusion map \cite{Coifman:2006cy}, where
eigenvectors of a finite dimensional matrix can be easily obtained
\textendash{} the eigenfunctions of the Laplace operator on a Riemannian
manifold are usually intractable. Fortunately, the high degree of
symmetry of a hypersphere allows the explicit construction of eigenfunctions,
called hyperspherical harmonics, via the projection of homogeneous
polynomials \cite{Atkinson:2012tz,Wen:1985fm}. The exact heat kernel
is then obtained as a convergent power series in these eigenfunctions.
In this paper, we compare the analytic behavior of this exact heat
kernel with that of the parametrix kernel and analyze their performance
in classification.

%
%
\section{Results}

The heat kernel is the fundamental solution to the heat equation
$(\partial_{t}-\Delta_{x})u(x,t)=0$ with an initial point source
\cite{GOLDBART:2016uz}, where $\Delta_{x}$ is the Laplace operator; 
the amount of heat emanating from the source
that has diffused to a neighborhood during time $t>0$ is
used to measure the similarity between the source and proximal
points. The heat conduction depends on the geometry of feature space,
and the main idea behind the application of hyperspherical geometry to
data analysis relies on the following  map from a
non-negative feature space to a unit hypersphere:
\begin{defn}
A hyperspherical  map $\varphi:\mathbb{R}_{\ge0}^{n}\setminus\{0\}\rightarrow S^{n-1}$
maps a vector $\mathbf{x}$, with $x_{i}\ge0$ and $\sum_{i=1}^{n}x_{i}>0$,
to a unit vector $\hat{x}\in S^{n-1}$ where $(\hat{x})_{i}\equiv\sqrt{x_{i}/\sum_{j=1}^{n}x_{j}}$.
\end{defn}
We will henceforth denote the image of a feature vector $\mathbf{x}$
under the hyperspherical  map as $\hat{x}$.  The notion of
neighborhood requires a well-defined measurement of distance on the
hypersphere, which is naturally the great arc length \textendash{} the
geodesic on a hypersphere. Both parametrix approximation and exact
solution employ the great arc length, which is related to the
following definition of cosine similarity:
\begin{defn}
The generic cosine similarity between two feature vectors
$\mathbf{x},\mathbf{y}\in\mathbb{R}^{n}\setminus\{0\}$ is 
\[
\cos\theta\equiv\frac{\mathbf{x}\cdot\mathbf{y}}{\left\Vert
    \mathbf{x}\right\Vert \left\Vert \mathbf{y}\right\Vert },
\]
where $\left\Vert \cdot\right\Vert $ is the Euclidean $L^{2}$-norm,
and $\theta\in[0,\pi] $ is the great arc length on $S^{n-1}$.  For
unit vectors $\hat{x}=\varphi(\mathbf{x})$ and
$\hat{y}=\varphi(\mathbf{y})$ obtained from non-negative feature
vectors $\mathbf{x},\mathbf{y}\in\mathbb{R}_{\ge0}^{n}\setminus\{0\}$
via the hyperspherical  map, the cosine similarity reduces
to the dot product $\cos\theta=\hat{x}\cdot\hat{y}$; the
non-negativity of $\bf x$ and $\bf y$ guarantees that
$\theta\in[0,\pi/2]$ in this case.
\end{defn}

\subsection{Parametrix expansion}

The parametrix kernel $K^{{\rm prx}}$ previously used in the literature
is just a Gaussian RBF function with $\theta=\arccos\hat{x}\cdot\hat{y}$
as the radial distance \cite{Lafferty:2005uy}:
\begin{defn}
The parametrix kernel is a non-negative function 
\[
K^{{\rm prx}}(\hat{x},\hat{y};t)={\rm e}^{-\frac{\arccos^{2}\hat{x}\cdot\hat{y}}{4t}}={\rm e}^{-\frac{\theta^{2}}{4t}},
\]
defined for $t>0$ and attaining global maximum $1$ at $\theta=0$.
\end{defn}
\noindent Note that this kernel is assumed to be restricted to the positive orthant.
The normalization factor $(4\pi t)^{-\frac{n-1}{2}}$ is
numerically unstable as $t\downarrow0$ and complicates hyperparameter
tuning; as a global scaling factor of the kernel can be absorbed into
the misclassification $C$-parameter in SVM, this overall normalization
term is ignored in this paper. Importantly, the parametrix kernel
$K^{{\rm prx}}$ is merely the Gaussian multiplicative factor without
any asymptotic expansion terms in the full parametrix expansion
$G^{{\rm prx}}$ of the  heat kernel on a hypersphere
\cite{Lafferty:2005uy,Berger:wXGLCovN}, as described below.

The Laplace operator on manifold $\mathcal M$ equiped with a
Riemannian metric
$g_{\mu\nu}$  acts on a function $f$ that depends only on the geodesic
distance $r$ from a fixed point as
\begin{equation}\label{eq:radialLaplacian}
\Delta f(r)  =f''(r)+\left(\log\sqrt{g}\right)'f'(r),
\end{equation}
where $g\equiv \det(g_{\mu\nu})$ and $'$ denotes the radial
derivative. Due to the nonvanishing metric derivative in
Equation~\ref{eq:radialLaplacian}, the canonical diffusion function
\begin{equation}
G(r,t)=\left(\frac{1}{4\pi t}\right)^{\frac{d}{2}}\exp\left(-\frac{r^{2}}{4t}\right) \label{eq:gaussianrbf}
\end{equation}
does not satisfy the heat equation; that is,
$(\Delta-\partial_t)G(r,t) \neq 0$ (Supplementary Material, Section
S2).  For sufficiently small time $t$ and geodesic distance $r$, the
parametrix expansion of the heat kernel on a full hypersphere proposes an approximate
solution
\[
K_p(r,t)=G(r,t)\left(u_{0}(r)+u_{1}(r)t+u_{2}(r)t^{2}+\cdots + u_p(r) t^p\right),
\]
where the functions $u_i$ should be found such that $K_p$ satisfies
the heat equation to order $t^{p-d/2}$, which is small for $t\ll1$ and
$p>d/2$; more precisely, we seek $u_i$ such that
\begin{equation}
(\Delta - \partial_t) K_p = G \,t^p\, \Delta u_p. \label{eq:parametrix}
\end{equation}

Taking the time derivative of $K_p$ yields 
\[
\partial_{t}K_p=G\cdot\left[\left(-\frac{d}{2t}+\frac{r^{2}}{4t^{2}}\right)\left(u_{0}+u_{1}t+u_{2}t^{2}+\cdots+u_pt^p\right)+\left(u_{1}+2u_{2}t+\cdots+pu_pt^{p-1}\right)\right],
\]
while the Laplacian of $K_p$ is
\[
\Delta K_p=\left(u_{0}+u_{1}t+\cdots+u_pt^p\right)\Delta G+G\Delta\left(u_{0}+u_{1}t+\cdots+u_pt^p\right)+2G'\left(u_{0}+u_{1}t+\cdots+u_pt^p\right)'.
\]

One can easily compute
\[
\Delta G=\left[\left(-\frac{1}{2t}+\frac{r^{2}}{4t^{2}}\right)-\frac{r}{2t}(\log\sqrt{g})'\right]G
\]
and
\[
G'\left(u_{0}+u_{1}t+\cdots\right)'=-\frac{r}{2t}\left(u_{0}'+u_{1}'t+\cdots\right)G.
\]

The left-hand side of Equation~\ref{eq:parametrix} is thus equal to
$G$ multiplied by
\begin{align*}
\left(u_{0}+\cdots+u_pt^p\right)\left[-\frac{r}{2t}(\log\sqrt{g})'+\frac{d-1}{2t}\right]+
\Delta\left(u_{0}+\cdots+u_pt^p\right)+&\\
-\frac{r}{t}\left(u_{0}'+\cdots+u_p't^p\right)-\left(u_{1}+2u_{2}t+\cdots+pu_pt^{p-1}\right)&,
\end{align*}
and we need to solve for $u_i$ such that all the coefficients of
$t^q$ in this expression, for $q<p$, vanish.

For $q=-1$, we need to solve 
\[
u_{0}\frac{r}{2}\left[-(\log\sqrt{g})'+\frac{d-1}{r}\right]=ru'_{0} \ ,
\]
or equivalently,
\[
\left(\log u_{0}\right)'=-\frac{1}{2}(\log\sqrt{g})'+\frac{d-1}{2r}.
\]
Integrating with respect to $r$ yields
\[
\log u_{0}=-\frac{1}{2}\left[\log\sqrt{g}-(d-1)\log r\right]+{\rm const.},
\]
where we implicitly take only the radial part of $\log \sqrt{g}$.
Thus, we get
\[
u_{0}={\rm const.}\times\left(\frac{\sqrt{g}}{r^{d-1}}\right)^{-\frac{1}{2}} \propto \left(\frac{\sin r}{r}\right)^{-\frac{d-1}{2}}
\]
as the zeroth-order term in the parametrix expansion.
Using this expression of $u_{0}$, the remaining terms become

\[
r\left[\left(u_{1}+u_{2}t+\cdots\right)(\log u_{0})'-\left(u_{1}'+u_{2}'t+\cdots\right)\right]+
\]
\[
+\left(\Delta u_{0}+t\Delta u_{1}+\cdots\right)-\left(u_{1}+2u_{2}t+\cdots\right),
\]
and we obtain the recursion relation
\[
u_{k+1}(\log u_{0})'-u_{k+1}'=-\frac{\Delta u_{k}-(k+1)u_{k+1}}{r}.
\]
Algebraic manipulations show that
\[
(\log r^{k+1}-\log u_{0}+\log u_{k+1})'u_{k+1}=r^{-1}\Delta u_{k}\, ,
\]
from which we get
\[
\left(\frac{u_{k+1}r^{k+1}}{u_{0}}\right)'=r^{(k+1)-1}u_{0}^{-1}\Delta
u_{k}.
\]
Integrating this equation and rearranging terms, we finally get
\begin{equation}
u_{k+1}=r^{-(k+1)}u_{0}\int_{0}^{r}d\tilde{r}\:\tilde{r}^{k}u_{0}^{-1}\Delta u_{k}. \label{eq:u_k+1}
\end{equation}

Setting $k=0$ in this recursion equation, we find the second correction term to be
\begin{eqnarray*}
u_{1} & = & \frac{u_{0}}{r}\int_{0}^{r}d\tilde{r}\:u_{0}^{-1}\Delta u_{0}\\
 & = & \frac{u_{0}}{r}\int_{0}^{r}d\tilde{r}\:u_{0}^{-1}\left(u_{0}''+u_{0}'(\log\sqrt{g})'\right).
\end{eqnarray*}
From our previously obtained solution for $u_0$, we find 
\[
u_{0}'=\frac{1}{2}\left(\frac{d-1}{r}-\frac{g'}{2g}\right)u_{0}.
\]
and
$$
u_{0}''   =  \frac{1}{4}\left[\frac{(d-1)(d-3)}{r^{2}}-\frac{g'(d-1)}{gr}-\frac{g''}{g}+\frac{5}{4}\left(\frac{g'}{g}\right)^{2}\right]u_{0}.
$$
Substituting these expressions into the recursion relation for $u_1$ yields
$$
u_{1}  =  \frac{u_{0}}{4r}\int_{0}^{r}dr\left[\frac{(d-1)(d-3)}{r^{2}}-\frac{g''}{g}+\frac{3}{4}\left(\frac{g'}{g}\right)^{2}\right].
$$
For the hypersphere $S^{d}$, where $d \equiv n-1$ and $g={\rm const.}\times\sin^{2(d-1)}r$,
we have
\[
\frac{g'}{g}=\frac{2(d-1)}{\tan r}
\]
and
\[
\frac{g''}{g}=2(d-1)\left(\frac{2d-3}{\tan^{2}r}-1\right).
\]
Thus,
\begin{eqnarray}
u_{1}&=&\frac{u_{0}}{4r}\int_{0}^{r}d\tilde{r}\left[\frac{(d-1)(d-3)}{\tilde{r}^{2}}-
  (d-1) \left( \frac{d-3}{\tan^2 \tilde{r}} -2\right)
\right]\nonumber\\
&=& \frac{u_0 (d-1)}{4r^2} \left[ 3-d + (d-1) r^2 + (d-3) r \cot r\right].\label{eq: para_u1}
\end{eqnarray}
Notice that  $u_1(r)=0$ when $d=1$ and
$u_1(r)=u_0(r)$ when $d=3$.  For $d=2$, $u_1/u_0$ is an increasing function in
$r$ and diverges to $\infty$ at $r=\pi$.  By contrast, for $d>3$, $u_1/u_0$ is
a decreasing function in $r$ and diverges to $-\infty$ at
$r=\pi$;  $u_1/u_0$ is relatively
constant for $r<\pi$ and starts to decrease rapidly only near
$\pi$.
Therefore, the first order correction is not able to remove the
unphysical behavior near $r=0$ in high dimensions where, according to
the first order parametrix kernel, the surrounding area is hotter than the heat
source.

Next, we apply Equation \ref{eq:u_k+1} again to obtain $u_2$ as
\begin{eqnarray*}
u_{2} & = & \frac{u_{0}}{r^{2}}\int_{0}^{r}d\tilde{r}\:\tilde{r}u_{0}^{-1}\Delta u_{1}\\
 & = & \frac{u_{0}}{r^{2}}\int_{0}^{r}d\tilde{r}\:\tilde{r}u_{0}^{-1}\left(u_{1}''+u_{1}'(\log\sqrt{g})'\right).
\end{eqnarray*}
After some cumbersome algebraic manipulations, we find

\begin{eqnarray}
\frac{u_{2}}{u_0}&=&\frac{d-1}{32}\left[(d-3)^{3}+\frac{(d-3)(d-5)(d-7)}{r^{4}}-\frac{(d-3)^{2}(d-5)}{r^{3}\tan r}\right.\nonumber\\
&&\left.+\frac{2(d-1)^{2}(d-3)}{r\tan r}+\frac{(d+1)(d-3)(d-5)}{r^{2}\sin r}\right].\label{eq: para_u2}
\end{eqnarray}

\noindent Again, $d=1$ and $d=3$ are special dimensions, where $u_{2}(r)=0$ for $d=1$,
and $u_{2}(r)=u_{0}/2$ for $d=3$;  for other dimensions, $u_{2}(r)$ is
singular at both $r=0$ and $\pi$.
Note that on $S^1$, the metric in geodesic polar coordinate is $g_{11} = 1$, so all parametrix expansion
coefficients $u_k(r)$ must vanish identically, as we have explicitly shown above.

Thus, the full $G^{{\rm prx}}$ defined on a hypersphere, where the
geodesic distance $r$ is just the arc length $\theta$, suffers from
numerous problems.  The zeroth order correction term $u_0 =
(\sin\theta/\theta)^{-\frac{n-2}{2}}$ diverges at $\theta=\pi$; this
behavior is not a major problem if $\theta$ is restricted to the range
$[0,\frac{\pi}{2}]$. Moreover, $G^{{\rm prx}}$ is also unphysical as
$\theta\downarrow0$ when $(n-2)t>3$; this condition on dimension and
time is obtained by expanding ${\rm
  e}^{-\theta^{2}/4t}=1-\frac{\theta^{2}}{4t}+{\mathcal
  O}(\theta^{4})$ and
$(\sin\theta/\theta)^{-\frac{n-2}{2}}=1+\frac{\theta^{2}}{12}(n-2)+{\mathcal
  O}(\theta^{3})$, and noting that the leading order $\theta^{2}$ term
in the product of the two factors is a non-decreasing function of
distance $\theta$ when $\frac{n-2}{12}\ge\frac{1}{4t}$, corresponding
to the unphysical situation of nearby points being hotter than the
heat source itself.  As the feature dimension $n$ is typically very
large, the restriction $(n-2)t<3$ implies that we need to take the
diffusion time to be very small, thus making the similarity measure
captured by $G^{{\rm prx}}$ decay too fast away from each data point
for use in clustering applications.
In this work, we further computed the first and second order correction terms,
denoted $u_1$ and $u_2$ in Equation~\ref{eq: para_u1} and Equation~\ref{eq:
   para_u2}, respectively.
In high dimensions, the divergence of
$u_1/u_0$ and $u_2/u_0$ at $\theta=\pi$ is not a major problem, as we expect the
expansion to be valid only in the vicinity $\theta\downarrow0$; however, the divergence
of $u_2/u_0$ at $\theta=0$ (to $-\infty$ in high dimensions) is pathological, and thus, we truncate
our approximation to ${\mathcal  O}(t^2)$. Since $u_1(\theta)$ is not able to
correct the unphysical behavior of the parametrix kernel near $\theta=0$ in
high dimensions, 
we conclude that the parametrix approximation fails in high dimensions.
Hence, the
only remaining part of $G^{{\rm prx}}$ still applicable to SVM
classification is the Gaussian factor, which is clearly not a heat
kernel on the hypersphere. The failure of this perturbative expansion
using the Euclidean heat kernel as a starting point suggests that
diffusion in $\mathbb{R}^{d}$ and $S^{d}$ are fundamentally different
and that the exact hyperspherical heat kernel derived from a
non-perturbative approach will likely yield better insights into the
diffusion process.

\subsection{Exact hyperspherical heat kernel}
By definition, the exact heat kernel $G^{{\rm ext}}(\hat{x},\hat{y};t)$
is the fundamental solution to heat equation $\partial_{t}u+\hat{L}^{2}u=0$
where $-\hat{L}^{2}$ is the hyperspherical Laplacian \cite{GOLDBART:2016uz,Grigoryan:1999du,Hsu:2002vq,Varopoulos:1987vx}.
In the language of operator theory, $G^{{\rm ext}}(\hat{x},\hat{y};t)$
is an integral kernel, or Green's function, for the operator $\exp\{-\hat{L}^{2}t\}$
and has an associated eigenfunction expansion. Because $\hat{L}^{2}$
and $\exp\{-\hat{L}^{2}t\}$ share the same eigenfunctions, obtaining
the eigenfunction expansion of $G^{{\rm ext}}(\hat{x},\hat{y};t)$
amounts to solving for the complete basis of eigenfunctions of $\hat{L}^{2}$.
The spectral decomposition of the Laplacian is in turn facilitated
by embedding $S^{n-1}$ in $\mathbb{R}^{n}$ and utilizing the global
rotational symmetry of $S^{n-1}$ in $\mathbb{R}^{n}$. The Euclidean
space harmonic functions, which are the solutions to the Laplace
equation $\nabla^{2}u=0$ in $\mathbb{R}^{n}$, can be projected to
the unit hypersphere $S^{n-1}$ through the usual separation of radial
and angular variables \cite{Atkinson:2012tz,Wen:1985fm}. In this
formalism, the hyperspherical Laplacian $-\hat{L}^{2}$ on $S^{n-1}$
naturally arises as the angular part of the Euclidean Laplacian on
$\mathbb{R}^{n}$, and $\hat{L}^{2}$ can be interpreted as the squared
angular momentum operator in $\mathbb{R}^{n}$ \cite{Wen:1985fm}.

The resulting eigenfunctions of $\hat{L}^{2}$ are known as the hyperspherical
harmonics and generalize the usual spherical harmonics in $\mathbb{R}^{3}$
to higher dimensions. Each hyperspherical harmonic is equipped with
a triplet of parameters or ``quantum numbers'' $(\ell,\{m_{i}\},\alpha)$:
the degree $\ell$, magnetic quantum numbers $\{m_{i}\}$ and $\alpha=\frac{n}{2}-1$.
In the eigenfunction expansion of $\exp\{-\hat{L}^{2}t\}$, we use
the addition theorem of hyperspherical harmonics to sum over the magnetic
quantum number $\{m_{i}\}$ and obtain the following main result: 
\begin{thm}
\label{thm:The-exact-hyperspherical-kernel-expansion}The exact hyperspherical
heat kernel $G^{{\rm ext}}(\hat{x},\hat{y};t)$ can be expanded as
a uniformly and absolutely convergent power series 
\[
G^{{\rm ext}}(\hat{x},\hat{y};t)=\sum_{\ell=0}^{\infty}{\rm
  e}^{-\ell(\ell+n-2)t}\frac{2\ell+n-2}{n-2}\frac{1}{A_{S^{n-1}}}C_{\ell}^{\frac{n}{2}-1}(\hat{x}\cdot\hat{y})
\]
 in the interval $\hat{x}\cdot\hat{y}\in[-1,1]$ and for $t>0$, where
$C_{\ell}^{\alpha}(w)$ are the Gegenbauer polynomials and $A_{S^{n-1}}=\frac{2\pi^{\frac{n}{2}}}{\Gamma\left(\frac{n}{2}\right)}$
is the surface area of $S^{n-1}$. Since the kernel depends on $\hat{x}$
and $\hat{y}$ only through $\hat{x}\cdot\hat{y}$, we will write
$G^{{\rm ext}}(\hat{x},\hat{y};t)=G^{{\rm ext}}(\hat{x}\cdot\hat{y};t)$. 
\end{thm}
\noindent\emph{Proof.} 
We will obtain  an eigenfunction expansion of the exact heat
  kernel by using the lemmas proved in Supplementary Material Section S2.5.3.
The completeness of hyperspherical harmonics (Lemma 1) states that
\begin{equation}\label{eq:completeness}
\delta(\hat{x},\hat{y})=\sum_{\ell=0}^{\infty}\sum_{\{m\}}Y_{\ell\{m\}}(\hat{x})Y_{\ell\{m\}}^{*}(\hat{y}).
\end{equation}
Applying the addition theorem (Lemma 2) to Equation~\ref{eq:completeness},
we get
$$
\delta(\hat{x},\hat{y})=\frac{1}{A_{S^{n-1}}}\sum_{\ell=0}^{\infty}\frac{2\ell+n-2}{n-2}C_{\ell}^{\frac{n}{2}-1}(\hat{x}\cdot\hat{y}).
$$
Next, we apply time evolution operator $e^{-t\hat{L}^2}$ on this
initial state to generate the
heat kernel
\begin{align}
G(\hat{x}\cdot\hat{y};t) & ={\rm e}^{-\hat{L}^{2}t}\delta(\hat{x},\hat{y})\\
 & =\sum_{\ell=0}^{\infty}{\rm e}^{-\ell(\ell+n-2)t}\frac{2\ell+n-2}{n-2}\frac{1}{A_{S^{n-1}}}C_{\ell}^{\frac{n}{2}-1}(\hat{x}\cdot\hat{y}).
\end{align}

To show that it is a uniformly and absolutely convergent
series for $t>0$, note that 
$$
|G(w;t)|\le\frac{1}{(n-2)A_{S^{n-1}}}\sum_{\ell=0}^{\infty}{\rm e}^{-\ell(\ell+n-2)t}(2\ell+n-2)\left|C_{\ell}^{\frac{n-2}{2}}(w)\right|,
$$
where $w=\hat{x}\cdot\hat{y}$. 

The  terms involving Gegenbauer
polynomials can be bounded by using  Lemma 3
as

\begin{align*}
\left|C_{\ell}^{\frac{n-2}{2}}(w)\right| & \le\left[w^{2}\frac{\Gamma(\ell+n-2)}{\Gamma(n-2)\Gamma(\ell+1)}+(1-w^{2})\frac{\Gamma(\frac{\ell+n-2}{2})}{\Gamma(\frac{n-2}{2})\Gamma(\frac{\ell}{2}+1)}\right]\\
 & =\left[\frac{\Gamma(\frac{\ell+n-2}{2})}{\Gamma(\frac{n-2}{2})\Gamma(\frac{\ell}{2}+1)}+\left(\frac{\Gamma(\ell+n-2)}{\Gamma(n-2)\Gamma(\ell+1)}-\frac{\Gamma(\frac{\ell+n-2}{2})}{\Gamma(\frac{n-2}{2})\Gamma(\frac{\ell}{2}+1)}\right)w^{2}\right]\\
 & \le\frac{\Gamma(\frac{\ell+n-2}{2})}{\Gamma(\frac{n-2}{2})\Gamma(\frac{\ell}{2}+1)}+\left|\frac{\Gamma(\ell+n-2)}{\Gamma(n-2)\Gamma(\ell+1)}-\frac{\Gamma(\frac{\ell+n-2}{2})}{\Gamma(\frac{n-2}{2})\Gamma(\frac{\ell}{2}+1)}\right|w^{2}\\
 & \le\frac{\Gamma(\frac{\ell+n-2}{2})}{\Gamma(\frac{n-2}{2})\Gamma(\frac{\ell}{2}+1)}+\left|\frac{\Gamma(\ell+n-2)}{\Gamma(n-2)\Gamma(\ell+1)}-\frac{\Gamma(\frac{\ell+n-2}{2})}{\Gamma(\frac{n-2}{2})\Gamma(\frac{\ell}{2}+1)}\right|\\
 & \equiv M_{\ell}.
\end{align*}

We thus have
\begin{align*}
|G(w;t)| & \le\frac{1}{(n-2)A_{S^{n-1}}}\sum_{\ell=0}^{\infty}{\rm e}^{-\ell(\ell+n-2)t}(2\ell+n-2)\left|C_{\ell}^{\frac{n-2}{2}}(w)\right|\\
 & \le\frac{1}{(n-2)A_{S^{n-1}}}\sum_{\ell=0}^{\infty}{\rm e}^{-\ell(\ell+n-2)t}(2\ell+n-2)M_{\ell}\\
 & \equiv\frac{1}{(n-2)A_{S^{n-1}}}\sum_{\ell=0}^{\infty}Q_{\ell}.
\end{align*}
But, in the large $\ell$ limit, the asymptotic expansion
$$
M_{\ell}\sim \frac{\ell^{n-3}}{(n-3)!}\,
$$
implies that
$$
\lim_{\ell\rightarrow\infty}\frac{Q_{\ell+1}}{Q_{\ell}}=\lim_{\ell\rightarrow\infty}\frac{{\rm
    e}^{-(2\ell+n-1)t}(2\ell+n)M_{\ell+1}}{(2\ell+n-2) M_{\ell}}= 0<1,
$$
for any $t>0$.
The sequence $\{Q_{\ell}\}$ is thus convergent, and hence, the
Weiestrass M-test implies that the eigenfunction expansion of the heat
kernel is uniformly and absolutely convergent in the indicated
intervals. {\hfill Q.E.D.}

Note that the exact kernel $G^{{\rm ext}}$ is a Mercer kernel
re-expressed by summing over the degenerate eigenstates indexed by $\{ m\}$.
As before, we will rescale the kernel by self-similarity and define: 
\begin{defn}
The exact kernel $K^{{\rm ext}}(\hat{x},\hat{y};t)$ is the exact
heat kernel normalized by self-similarity:
\[
K^{{\rm ext}}(\hat{x},\hat{y};t)=\frac{G^{{\rm ext}}(\hat{x}\cdot\hat{y};t)}{G^{{\rm ext}}(1;t)},
\]
which is defined for $t>0$, is non-negative, and attains global maximum
$1$ at $\hat{x}\cdot\hat{y}=1$. 
\end{defn}

\begin{figure}[t]
\begin{centering}
\includegraphics[width=6in]{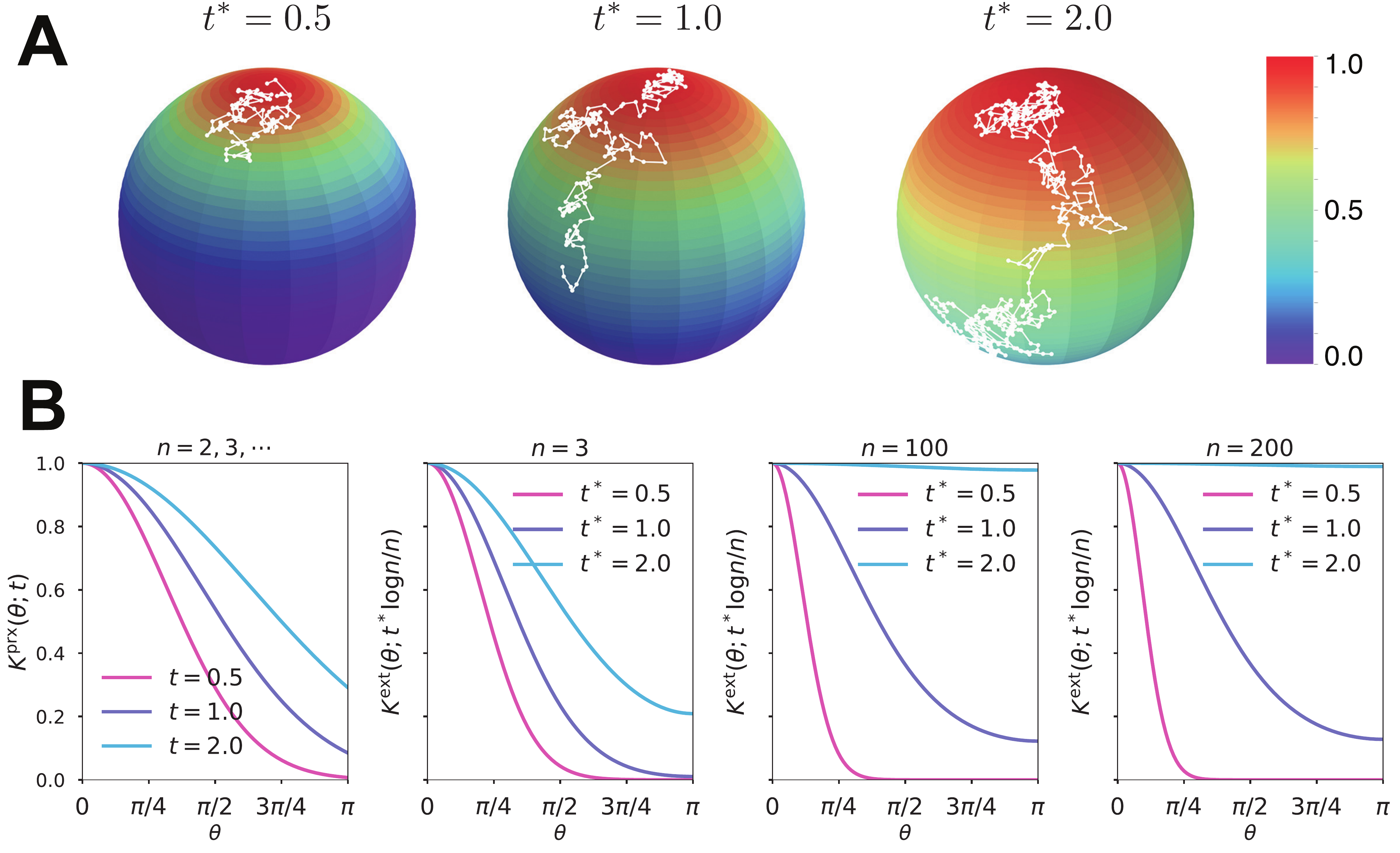}
\par\end{centering}
\caption{\label{fig:HK-randomwalks}(A) Color maps of the exact kernel $K^{{\rm ext}}$
on  $S^{2}$ at rescaled time $t^{*}=0.5,1.0,2.0$; the white paths are simulated random walks on $S^{2}$
with the Monte Carlo time approximately equal to $t=t^{*}\log3/3$. (B)
Plots of the parametrix kernel $K^{{\rm prx}}$ and exact kernel $K^{{\rm ext}}$
on  $S^{n-1}$, for  $n=3,100,200$,
as functions of the geodesic distance.}
\end{figure}

Note that unlike $K^{{\rm prx}}(\hat{x},\hat{y};t)$, $K^{{\rm ext}}(\hat{x},\hat{y};t)$ explicitly depends on the feature dimension $n$.
In general, SVM kernel hyperparameter tuning can be computationally costly
for a data set with both high feature dimension and large sample size.
In particular, choosing an appropriate diffusion time scale is an
important challenge. On the one hand, choosing a very large value
of $t$ will make the series converge rapidly; but, then, all points
will become uniformly similar, and the kernel will not be very useful.
On the other hand, a too small value of $t$ will make most data pairs
too dissimilar, again limiting the applicability of the kernel. In
practice, we thus need a guideline for a finite time scale at which
the degree of ``self-relatedness'' is not singular, but still larger
than the ``relatedness'' averaged over the whole hypersphere. Examining
the asymptotic behavior of the exact heat kernel in high feature dimension
$n$ shows that an appropriate time scale is $t\sim{\mathcal  O}(\log n/n)$;
in this regime the numerical sum in Theorem \ref{thm:The-exact-hyperspherical-kernel-expansion}
satisfies a stopping condition at low orders in $\ell$ and the sample points are in
moderate diffusion proximity to each other so that they can be accurately
classified (Supplementary Material, Section S2.5.4).

Figure~\ref{fig:HK-randomwalks}A illustrates the diffusion process
captured by our exact  kernel $K^{{\rm ext}}(\hat{x},\hat{y};t)$ in
three feature dimensions at time $t=t^{*}\log 3/3$, for
$t^{*}=0.5,1.0, 2.0$. In Figure~\ref{fig:HK-randomwalks}B, we systematically compared
the behavior 
of (1) dimension-independent parametrix kernel $K^{{\rm prx}}$ at
time $t=0.5, 1.0,2.0$ and (2) exact kernel $K^{{\rm ext}}$ on $S^{n-1}$
at $t=t^{*}\log n/n$ for $t^{*}=0.5,1.0,2.0$ and  $n=3,100,200$.
By symmetry, the slope of $K^{{\rm ext}}$
vanished at the south pole $\theta=\pi$ for any time $t$ and dimension
$n$. In sharp contrast, $K^{{\rm prx}}$ had a negative slope at $\theta=\pi$, again highlighting a singular behavior of the parametrix kernel. The ``relatedness'' measured by $K^{{\rm ext}}$ at the sweet
spot $t=\log n/n$ was finite over the whole hypersphere with
sufficient contrast between nearby and far away points. Moreover, the characteristic behavior of $K^{{\rm ext}}$ at $t=\log n/n$
did not change significantly for different values of the feature dimension $n$, confirming that the optimal
$t$ for many classification applications will likely reside near the ``sweet spot'' $t=\log n/n$.


\subsection{SVM classifications}
Linear SVM seeks a separating hyperplane that maximizes the margin,
i.e. the distance to the nearest data point. The primal formulation of
SVM attempts to minimize the norm of the weight vector $\mathbf{w}$
that is normal to the separating hyperplane, subject to either hard or
soft margin constraints. In the so-called Lagrange dual formulation of
SVM, one applies the Representer Theorem to rewrite the weight as a
linear combination of data points; in this set-up, the dot products of
data points naturally appear, and kernel SVM replaces the dot product
operation with a chosen kernel evaluation. The ultimate hope is that
the data points will become linearly separable in the new feature
space implicitly defined by the kernel.

We evaluated the performance of kernel SVM using the
\begin{enumerate}
\item linear kernel $K^{{\rm lin}}(\mathbf{x},\mathbf{y})=\mathbf{x}\cdot\mathbf{y}$,
\item Gaussian RBF $K^{{\rm rbf}}(\mathbf{x},\mathbf{y};\gamma)=\exp\{-\gamma|\mathbf{x}-\mathbf{y}|^{2}\}$, 
\item cosine kernel $K^{{\rm cos}}(\hat{x},\hat{y})=\hat{x}\cdot\hat{y}$, 
\item parametrix kernel $K^{{\rm prx}}(\hat{x},\hat{y};t)$, and 
\item exact kernel $K^{{\rm ext}}(\hat{x},\hat{y};t)$,
\end{enumerate}
on two independent data sets: (1) WebKB data of websites from four
universities (WebKB-4-University) \cite{Craven:1998tq}, and (2)
glioblastoma multiforme (GBM) mutation data from The Cancer Genome
Atlas (TCGA) with 5-fold cross-validations (CV)
(Supplementary Material, Section S1). 
The WebKB-4-University data contained 4199 documents in
total comprising four classes: student (1641), faculty (1124), course
(930), and project (504); in our analysis, however, we selected an
equal number of representative samples from each class, so that the
training and testing sets had balanced classes. Table~\ref{tab:WebKB} shows the
average optimal prediction accuracy scores of the five kernels for a
varying number of representative samples, using 393 most frequent word
features (Supplementary Material, Section S1). 
The exact kernel outperformed the
Gaussian RBF and parametrix kernel, reducing the error by
$41\%\sim45\%$ and by $1\%\sim7\%$, respectively.
Changing the feature dimension did not affect the performance much
(Table \ref{tab:SI-WebKB-4-University}).

\begin{table}

\centering

\begin{tabular}{lccccc}
\hline\hline
$m_{{\rm r}}$ & lin & rbf & cos & prx & ext\\
\hline 
100 & 74.2\% & 75.1\% & 84.4\% & 85.4\% & \textbf{85.6\%}\\

200 & 80.9\% & 82.0\% & 89.2\% & 89.6\% &  \textbf{89.9\%}\\

300 & 83.2\% & 84.1\% & 89.9\% & 90.5\% &  \textbf{91.1\%}\\

400 & 86.7\% & 86.1\% & 91.3\% & 91.7\% & \textbf{92.3\%}\\
\hline\hline 

\end{tabular}

\caption{
\label{tab:WebKB}
WebKB-4-University Document Classification. 
Performance test on four-class (\emph{student}, \emph{faculty}, \emph{course}, and \emph{project})
classification of WebKB-4-University word count data with different
number $m_{{\rm r}}$ of representatives for each class, for $m_{{\rm r}}=100,200,300,400$.
The entries show the average of optimal 5-fold cross-validation mean
accuracy scores of five runs. The exact kernel (ext) reduced the error of
parametrix kernel (prx) by $1\%\sim7\%$ and the Gaussian RBF (rbf)
by $41\%\sim 45\%$; the cosine kernel (cos) also reduced the error
of linear kernel (lin) by $34\%\sim 43\%$.
        }
\end{table}
\begin{table}[htb]
\centering
\begin{tabular}{llccccc}
\hline\hline
$n$ & $m_{{\rm r}}$ & lin & rbf & cos & prx & ext\\\hline

393 & 400 & 86.73\% & 86.27\% & 91.57\% & 91.99\% &  \textbf{92.44\%}\\
726 & 400 & 86.78\% & 86.95\% & 92.62\% & 92.91\% &  \textbf{93.00\%}\\
1023 & 400 & 85.56\% & 86.11\% & 92.62\% & 92.74\% &  \textbf{92.91\%}\\
1312 & 400 & 85.78\% & 86.75\% & 92.56\% & 92.81\% &  \textbf{93.03\%}\\\hline\hline
\end{tabular}

\caption{\label{tab:SI-WebKB-4-University}WebKB-4-University Document Classification. Comparison of kernel SVMs on the WebKB-4-University data
  with a fixed sample size $m_{{\rm r}}$, but varying feature
  dimension $n$.  To account for the randomness in
  selecting the representative samples using KMeans
  (Supplementary Material, Section S1), 
  we performed fives runs of representative selection, and
  then performed CV using the training and test sets obtained from
  each run.
  Finally, we  averaged the five mean CV scores to assess the
  performance of each classifier on the imbalanced
  WebKB-4-University data set.  The exact (ext) and cosine (cos) 
  kernels outperformed the Gaussian RBF (rbf) and linear (lin) kernels
  in various feature dimensions $n=393,726,1023,$ and $1312$, with
  fixed and balanced class size $m_{{\rm r}}=400$. 
  A word was selected as a feature if its total count was greater than
  1/10, 1/20, 1/30 or 1/40 times the total number of web pages in the
  WebKB-4-University data set, with the different thresholds
  corresponding to the different rows in the table.
  The exact kernel reduced
  the errors of Gaussian RBF and parametrix kernels by $45\sim48\%$ and
  $1\sim6\%$, respectively; the cosine kernel reduced the errors of
  linear
  kernel by $36\sim49\%$.}
\end{table}

\begin{table}

\centering
\begin{tabular}{rccccc}
\hline \hline 
    & lin & rbf & cos & prx & ext \\
\hline
\textit{ZMYM4}   & 82.9\% & 84.0\% & 83.6\% & 84.1\% & \textbf{85.1\%}\\
\textit{ADGRB3}  & 75.7\% & \textbf{81.0\%} & 78.0\% & 79.5\% & 79.3\%\\
\textit{NFX1}    & 73.0\% & 81.2\% & 80.9\% & \textbf{82.7\%} & 82.5\%\\
\textit{P2RX7}   & 79.2\% & 84.1\% & \textbf{85.0\%} & 84.0\% & \textbf{85.0\%}\\
\textit{COL1A2}  & 68.4\% & 70.5\% & 72.9\% & 73.9\% & \textbf{74.2\%}\\
\hline \hline

\end{tabular}
\caption{\label{tab:TCGA}TCGA-GBM Genotype Imputation. Performance test on binary classification of \emph{mutant} vs.~\emph{wild-type} in TCGA-GBM
mutation count data. The rows are different genes, the mutation statuses
of which were imputed using  $m_{{\rm r}}$ samples in each mutant and
wild-type class. The entries show the average of optimal 5-fold cross-validation mean accuracy
scores of five runs.
}
\end{table}

In the TCGA-GBM data, there were 497 samples, and we aimed to impute
the mutation status of one gene -- i.e., mutant or wild-type -- from the mutation
counts of other genes. For each imputation target, we first counted
the number $m_{{\rm r}}$ of mutant samples and then selected an equal number of wild-type samples
for  5-fold CV. Imputation tests were performed for top 102 imputable
genes (Supplementary Material, Section S1). 
Table~\ref{tab:TCGA} shows the
average prediction accuracy scores for 5 biologically interesting
genes known to be important for cancer \cite{Hanahan:2011gu}:
\begin{enumerate}
\item \emph{ZMYM4} ($m_{{\rm r}}=33$) is implicated in an antiapoptotic activity;
\cite{Smedley:1999uq,Shchors:2002hu};
\item \emph{ADGRB3} ($m_{{\rm r}}=37$) is a brain-specific angiogenesis
inhibitor \cite{Zohrabian:2007wv,Kaur:2010jf,Hamann:2015hv};
\item \emph{NFX1} ($m_{{\rm r}}=42$) is a repressor of \emph{hTERT}
transcription \cite{Yamashita:2016dm} and is thought to regulate inflammatory
response \cite{Song:1994bh};
\item \emph{P2RX7} ($m_{{\rm r}}=48$) encodes an ATP receptor which plays
a key role in restricting tumor growth and metastases \cite{Adinolfi:2015kk,GomezVillafuertes:2015fd,LinanRico:2016gu};
\item \emph{COL1A2} ($m_{{\rm r}}=61$) is overexpressed in the medulloblastoma
microenvironment and is a potential therapeutic target \cite{Anderton:2008gh,Liang:2007kw,Schwalbe:2011ic}.
\end{enumerate}

\begin{figure}[ht]
\begin{centering}
\includegraphics[width=6in]{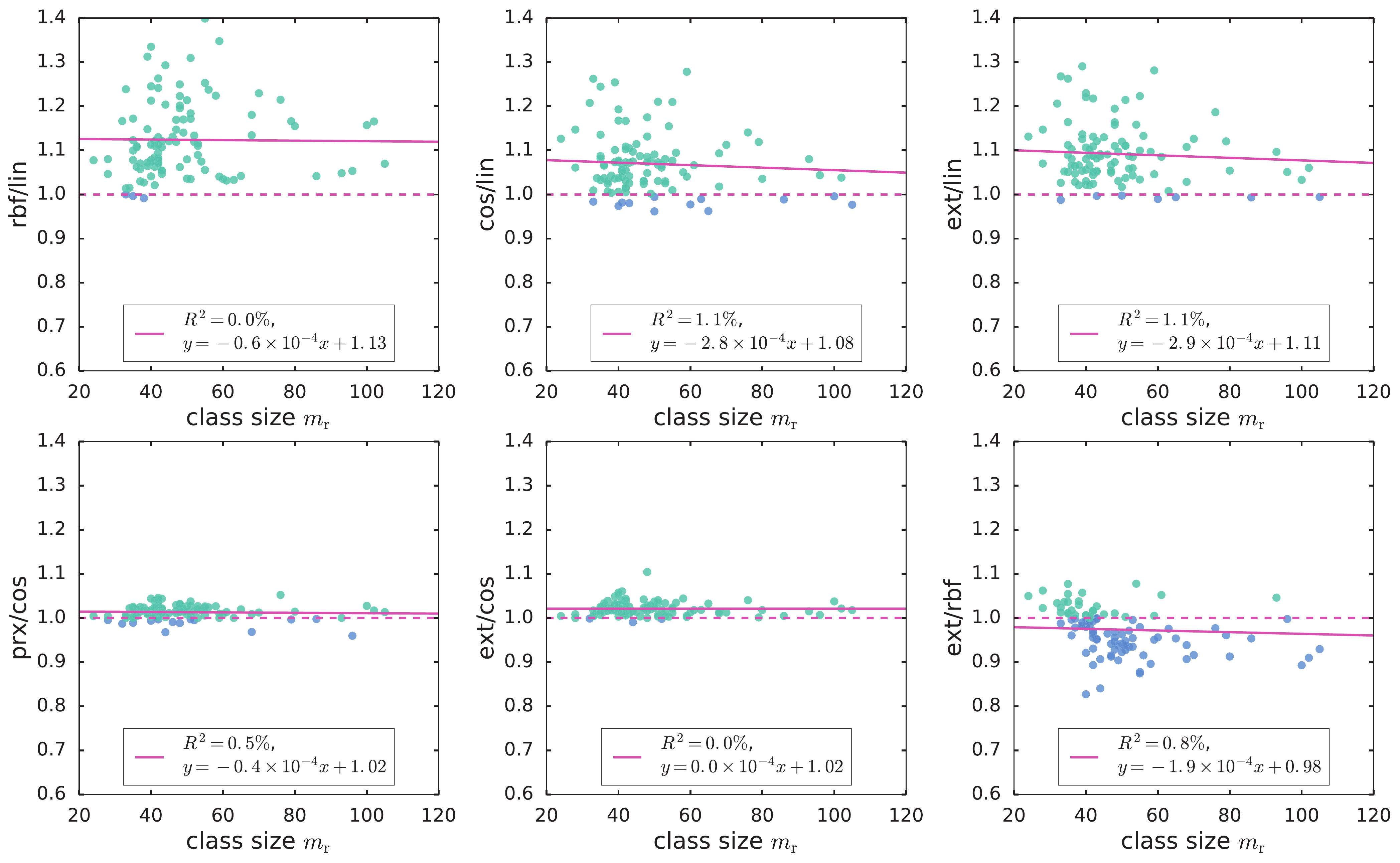}
\par\end{centering}
\caption{\label{fig:HK-SI-scores}Comparison of the classification
  accuracy of SVM using linear (lin), cosine (cos), Gaussian RBF
  (rbf), parametrix (prx), and exact (ext) kernels on TCGA mutation
  count data. The plots show the ratio of accuracy scores for two
  different kernels. For visualization purpose, we excluded one gene
  with $m_{{\rm r}}=250$. The ratios rbf/lin, prx/cos, and ext/cos
  were essentially constant in class size $m_{{\rm r}}$ and greater than 1; in other words, the Gaussian RBF (rbf)
  kernel outperformed the
  linear (lin) kernel, while the exact (ext) and parametrix (prx) kernels
  outperformed the cosine (cos) kernel uniformly over all values of class size
  $m_{{\rm r}}$.  However, the more negative slope in the linear fit
  of cos/lin  hints that the accuracy scores of
  cosine and linear kernels may depend on the class
  size $m_{{\rm r}}$; the exact kernel also tended to outperform Gaussian RBF kernel when $m_{\rm r}$ was small. }
\end{figure}

\begin{figure}[t]
\begin{centering}
\includegraphics[width=6in]{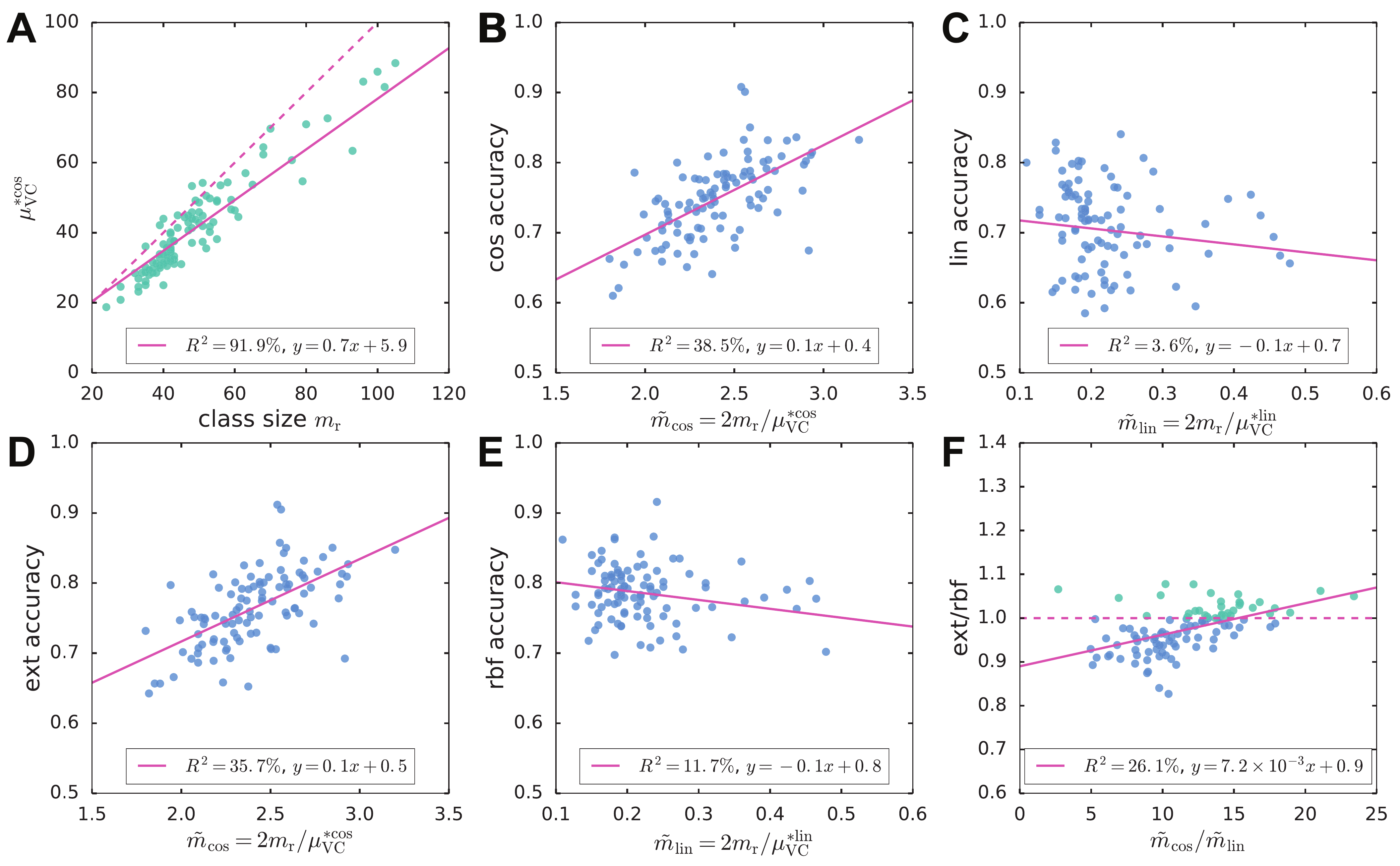}
\par\end{centering}
\caption{\label{fig:HK-SI-VC}(A) A strong linear relation is seen
  between the VC-bound for cosine kernel $\mu_{{\rm VC}}^{*\cos}$ and class
  size $m_{{\rm r}}$. The dashed line marks $y=x$; the VC-bound
  for linear kernel, however, was a constant $\mu_{{\rm VC}}^{*{\rm
      lin}}=439$. (B-E) The scatter plots of accuracy scores for
    cosine (cos), linear (lin), exact (ext), and Gaussian RBF (rbf)
    kernels vs. the effective sample size $\tilde{m} = 2m_{{\rm
        r}}/\mu_{{\rm VC}}^{*}$; the accuracy scores of exact and cosine
    kernels increased with the effective sample size, whereas those of
    Gaussian RBF and linear kernels tended to decrease with the effective
    sample size. (F) The ratio of ext vs. rbf accuracy scores is
    positively correlated with the ratio  $\tilde{m}_{\rm cos}/\tilde{m}_{\rm lin}$
of effective sample sizes.}
\end{figure}

For the remaining genes, the exact kernel generally outperformed the
linear, cosine and parametrix kernels (Figure~\ref{fig:HK-SI-scores}).  However, even though
the exact kernel dramatically outperformed the Gaussian RBF in the
WebKB-4-University classification problem, the advantage of the exact
kernel in this mutation analysis was not evident (Figure~\ref{fig:HK-SI-scores}). It is
possible that the radial degree of freedom $\sum_{i=1}^{n}x_{i}$ in
this case, corresponding to the genome-wide mutation load in each
sample, contained important covariate information not captured by the
hyperspherical heat kernel. The difference in accuracy between the
hyperspherical kernels (cos, prx, and ext) and the Euclidean kernels
(lin and rbf) also hinted some weak dependence on class size $m_{\rm
  r}$ (Figure~\ref{fig:HK-SI-scores}), or equivalently the sample size $m=2m_{\rm r}$. In
fact, the level of accuracy showed much stronger correlation with the
``effective sample size'' $\tilde{m}$ related to the empirical
Vapnik-Chervonenkis (VC) dimension
\cite{Vladimir:l-V4Juw7,Boser:1992fz,Guyon:1993ub,Vapnik:1994dk,Anonymous:2001vo}
of a kernel SVM classifier (Figure~\ref{fig:HK-SI-VC}A-E); moreover, 
the advantage of the exact kernel over the Guassian RBF kernel
grew with the effective sample size ratio $\tilde{m}_{\rm
  cos}/\tilde{m}_{\rm lin}$ (Figure~\ref{fig:HK-SI-VC}F, Supplementary Material, Section S2.5.5). 

By construction, our definition of the hyperspherical map exploits
only the positive portion of the whole hypersphere, where the
parametrix and exact heat kernels seem to have similar
performances. However, if we allow the data set to assume negative
values, i.e.~the feature space is the usual
$\mathbb{R}^{n}\backslash\{0\}$ instead of
$\mathbb{R}_{\ge0}^{n}\backslash\{0\}$, then we may apply the usual
projective map, where each vector in the Euclidean space is normalized
by its $L^{2}$-norm. As shown in
  Figure~\ref{fig:HK-randomwalks}B, the parametrix kernel is
  singular at $\theta=\pi$ and qualitatively deviates from the exact
  kernel for large values of $\theta$. Thus, when  data points
populate the whole hypersphere, we expect to find more significant
differences in performance between the exact and parametrix
kernels. For example, Table \ref{tab:SI-SP500} shows the kernel SVM classifications
of 91 S\&P500 \emph{Financials} stocks against 64 \emph{Information
  Technology} stocks ($m=155$) using their log-return instances
between January 5, 2015 and November 18, 2016 as features.  As long as
the number of features was greater than sample size, $n>m$, the exact
kernel outperformed all other kernels and reduced the error of
Gaussian RBF by $29\sim51\%$ and that of parametrix kernel by
$17\sim51\%$.

\begin{table}[htb]

\centering
\begin{tabular}{rrccccc}

\hline\hline
$n$ & $m$ & lin & rbf & cos & prx & ext\\\hline
475  & 155 & 98.06\% & 98.69\% & 98.69\% & 98.69\% &  \textbf{99.35\%}\\
238 & 155 & 95.50\% & 96.77\% & 94.82\% & 96.13\% &  \textbf{98.06\%}\\
159 & 155 & 94.86\% & 95.48\% & 95.48\% & 96.13\% &  \textbf{96.79\%}\\
119 & 155 & 92.86\% & 93.53\% & 91.57\% & \textbf{94.15\%} & \textbf{94.15\%}\\
95  & 155 & 91.55\% & \textbf{95.50\%} & 94.19\% & 94.15\% &  94.79\%\\\hline\hline

\end{tabular}

\caption{\label{tab:SI-SP500}S\&P500 Stock Classification. Classifications were performed on $m=155$ stocks from S\&P500 companies:
  91 \emph{Financial} vs.~64 \emph{Information Technology}. The 475
  log-return instances  between January 5, 2015 and November
  18, 2016 were used as features. We uniformly subsampled the
  instances to generate variations in the feature dimension $n$. Here, we
  report the mean 5-fold CV accuracy score for each kernel.
  Although the two classes were slightly imbalanced, all scores were much
  larger than the ``random score'' $91/155\approx58.7\%$, calculated
  from the
  majority class size and sample size. For $n>m$, the exact (ext) kernel outperformed all
  other kernels and reduced the errors of Gaussian RBF (rbf) and parametrix
  (prx) kernels by $29\sim51\%$ and $17\sim51\%$, respectively. When
  $n<m$, the exact kernel started to lose its advantage over the Gaussian RBF kernel.}

\end{table}

\section{Discussion}

This paper has constructed the exact hyperspherical heat kernel using
the complete basis of high-dimensional angular momentum eigenfunctions
and tested its performance in kernel SVM. We have shown that the exact
kernel and cosine kernel, both of which employ the hyperspherical
maps, often outperform the Gaussian RBF and linear kernels. The
advantage of using hyperspherical kernels likely arises from the
hyperspherical maps of feature space, and the exact kernel may
further improve the decision boundary flexibility of the raw cosine
kernel. To be specific, the hyperspherical maps remove the less
informative radial degree of freedom in a nonlinear fashion and
compactify the Euclidean feature space into a unit hypersphere where
all data points may then be enclosed within a finite radius.  By
contrast, our numerical estimations using TCGA-GBM data show that for
linear kernel SVM, the margin $M$ tends to be much smaller than the
data range $R$ in order to accommodate the separation of strongly
mixed data points of different class labels; as a result, the ratio
$R/M$ was much larger than that for cosine kernel SVM. This insight
may be summarized by the fact that the upper bound on the empirical
VC-dimension of linear kernel SVM tends to be much larger than that
for cosine kernel SVM, especially in high dimensions, suggesting that
the cosine kernel SVM is less sensitive to noise and more
generalizable to unseen data. The exact kernel is equipped with an
additional tunable hyperparameter, namely the diffusion time $t$,
which adjusts the curvature of nonlinear decision boundary and thus
adds to the advantage of hyperspherical maps. Moreover, the
hyperspherical kernels often have larger effective sample sizes than
their Euclidean counterparts and, thus, may be especially useful for
analyzing data with a small sample size  in high feature dimensions.

The failure of the parametrix expansion of heat kernel, especially in
dimensions $n\gg 3$, signals a dramatic difference between diffusion
in a non-compact space and that on a compact manifold.  It remains to
be examined how these differences in diffusion process, random walk
and topology between non-compact Euclidean spaces and compact
manifolds like a hypersphere help improve clustering performance as supported by the results of this paper.

\section*{Funding}

This research was supported by a Distinguished
  Scientist Award from Sontag Foundation and the Grainger Engineering
  Breakthroughs Initiative.

\section*{Acknowledgments}
We thank Alex Finnegan and Hu Jin for critical reading of the manuscript
  and helpful comments. We also thank Mohith Manjunath for his help with the TCGA data.

\newpage

\setcounter{figure}{0}
\setcounter{equation}{0}
\setcounter{section}{0}
\renewcommand{\thefigure}{S\arabic{figure}}
\renewcommand{\theequation}{S\arabic{equation}}
\renewcommand{\thesection}{S\arabic{section}}

%

\centerline{\LARGE\bf Supplementary Material}

\section{Data preparation and SVM classification}\label{appendix_data}

The WebKB-4-University raw webpage data were downloaded from \url{http://www.cs.cmu.edu/afs/cs/project/theo-20/www/data/} and
processed with the python packages Beautiful Soup and Natural Language
Toolkit (NLTK). Our feature extraction excluded punctuation marks and
included only letters and numerals where capital letters were all
converted to lower case and each individual digit 0-9 was represented
by a ``\#.'' Very infrequent words, such as misspelled words,
non-English words, and words mixed with special characters, were
filtered out.  We selected top $393$ most frequent words as features
in our classification tests; the cutoff was chosen to select frequent
words whose counts across all webpage documents are greater than
$10\%$ of the total number of documents. There were 4199 documents in
total: student (1641), faculty (1124), course (930), and project
(504).

The TCGA-GBM data were downloaded from the GDC Data Portal under the
name TCGA-GBM Aggregated Somatic Mutation. The mutation count data set
was extracted from the MAF file, while ignoring the detailed types of
mutations and counting only the total number of mutations in each
gene. Very infrequently, mutated genes were filtered out if the total
number of mutations in one gene across all samples is less than $10\%$
of the total number of samples ($m=497$ samples and $n=439$ genes). We
imputed the mutation status of one gene, mutant or wild-type, from the
mutation counts of the remaining genes. The most imputable genes were
selected using 5-fold cross-validation linear kernel SVM. Most of the
mutant and wild-type samples were highly unbalanced, the ratio being
typically around $1:9$; therefore, unthresholded area-under-the-curve
(AUC) of the receiver operating characteristic (ROC) curve was used to
quantify the classification performance of the linear kernel
SVM. Mutated genes with AUC greater than $60\%$ were selected for the
subsequent imputation tests.

To balance the sample size between classes, we performed K-means
clustering of samples within each class, with a specified number
$m_{{\rm r}}$ of centroids and took the samples closest to each
centroid as representatives. For the WebKB document classifications,
we used $m_{{\rm r}}\le\min\{m_{{\rm student }},m_{{\rm
    faculty}},m_{{\rm course }},$ $m_{\rm project}\}$, and
K-means clustering was performed in each of the four classes
separately; for the TCGA-GBM data, $m_{{\rm r}}$ was chosen to be the
number of samples in each mutant (minority) class, and K-means
clustering was performed in the wild-type (majority) class. Since
K-means might depend on the random initialization, we performed the
clustering 50 times and selected the top $m_{{\rm r}}$ most frequent
representatives. Five-fold stratified cross-validations (CV) were
performed on the resulting balanced data sets, where training and test
samples were drawn without replacement from each class. The mean CV
accuracy scores across the five folds were recorded. 



\section{Hyperspherical Heat Kernel}

\subsection{Laplacian on a Riemannian manifold}

The Laplacian on a Riemannian manifold ${\mathcal M}$ with metric $g_{\mu\nu}$
is the operator
$$
\Delta: C^{\infty}({\mathcal M}) \rightarrow C^{\infty}({\mathcal M})
$$
 defined as
\begin{equation}\label{eq:Laplacian}
\Delta\equiv\frac{1}{\sqrt{g}}\partial_{\mu}\left(\sqrt{g}g^{\mu\nu}\partial_{\nu}\right), 
\end{equation}
where $g=|\det g|$, and the Einstein summation convention is used.
It can be also written in terms of the covariant derivative $\nabla_{\mu}$ as
\begin{equation}\label{eq:Laplacian2}
\Delta=g^{\mu\nu}\nabla_{\mu}\nabla_{\nu}.
\end{equation}
The covariant derivative satisfies the following
properties
$$
\nabla_{\mu}f=\partial_{\mu}f,\quad f\in C^{\infty}({\mathcal M})
$$
$$
\nabla_{\mu}V^{\nu}=\partial_{\mu}V^{\nu}+\Gamma_{\:\lambda\mu}^{\nu}V^{\lambda},\quad V\in T_{p}{\mathcal M}
$$
$$
\nabla_{\mu}\omega_{\nu}=\partial_{\mu}\omega_{\nu}-\Gamma_{\:\nu\mu}^{\lambda}\omega_{\lambda},\quad\omega\in T_{p}^{*}{\mathcal M},
$$
where $\Gamma_{\:\alpha\beta}^{\lambda}$ is the Levi-Civita connection
satisfying
$\Gamma_{\:\alpha\beta}^{\lambda}=\Gamma_{\:\beta\alpha}^{\lambda}$
and $\nabla_{\lambda}g_{\mu\nu}=0$.
To show Equation~\ref{eq:Laplacian2},
recall that the Levi-Civita connection is uniquely
determined by the geometry, or the metric tensor, as
$$
\Gamma_{\:\alpha\beta}^{\lambda}=\frac{1}{2}g^{\lambda\rho}\left(\partial_{\alpha}g_{\beta\rho}+\partial_{\beta}g_{\alpha\rho}-\partial_{\rho}g_{\alpha\beta}\right).
$$
Using the formula for determinant
differentiation
$$
\left[\log\left(\det\mathbf{A}\right)\right]'={\rm tr}\left(\mathbf{A}'\mathbf{A}^{-1}\right),
$$
we can thus write
$$
\Gamma_{\:\lambda\mu}^{\lambda} =\partial_{\mu}\log\sqrt{g}.
$$
Hence, for any $f\in C^{\infty}({\mathcal M})$,
\begin{align*}
 g^{\mu\nu}\nabla_{\mu}\nabla_{\nu}f & = \nabla_{\mu}(g^{\mu\nu}\partial_{\nu}f)\\
 &=\partial_{\mu}(g^{\mu\nu}\partial_{\nu}f) + \Gamma_{\:\lambda\mu}^{\lambda} (g^{\mu\nu}\partial_{\nu}f)\\
 &=\partial_{\mu}(g^{\mu\nu}\partial_{\nu}f) + (\partial_{\mu}\log\sqrt{g} )(g^{\mu\nu}\partial_{\nu}f)\\
 & =\frac{1}{\sqrt{g}}\partial_{\mu}\left(\sqrt{g}g^{\mu\nu}\partial_{\nu}\right),
\end{align*}

proving the equivalence of Equation~\ref{eq:Laplacian} and Equation~\ref{eq:Laplacian2}.

\subsection{The induced metric on $S^{n-1}$}

The $(n-1)$-sphere  embedded in $\mathbb{R}^{n}$ can be parameterized as
\begin{eqnarray*}
x_{1} & = & \cos\theta_{1}\\
x_{2} & = & \sin\theta_{1}\cos\theta_{2}\\
x_{3} & = & \sin\theta_{1}\sin\theta_{2}\cos\theta_{3}\\
 & \vdots  & \\
x_{n-1} & = & \sin\theta_{1}\cdots\sin\theta_{n-2}\cos\theta_{n-1}\\
x_{n} & = & \sin\theta_{1}\cdots\sin\theta_{n-2}\sin\theta_{n-1},
\end{eqnarray*}
where $0\leq\theta_i\leq\pi$, for $i=1,\ldots,n-2$, and  $0\leq\theta_{n-1}\leq2\pi$.

Let $\lambda:=(\partial x_{i}/\partial\theta_{j})$ denote
the $n\times(n-1)$ Jacobian matrix for the above
coordinate transformation.  
The square of the line element in $\mathbb{R}^{n}$ is given by 
$$
ds_{n}^{2}=\sum_{i=1}^{n}dx_idx_i.
$$
Restricted to $S^{n-1}$,
$$
dx_i=\sum_{j=1}^{n-1}\frac{\partial x_{i}}{\partial\theta_{j}}d\theta_{j}=\sum_{j=1}^{n-1}\lambda_{ij}d\theta_{j}.
$$
Therefore, on $S^{n-1}$, we have
\begin{align*}
ds_{n-1}^{2} & =\sum_{i=1}^{n}\sum_{j,j'=1}^{n-1}\lambda_{ij}\lambda_{ij'}d\theta_{j}d\theta_{j'}\\
 &
 =\sum_{j,j'=1}^{n-1}\left(\sum_{i=1}^{n}\lambda_{ij}\lambda_{ij'}\right)d\theta_{j}
 d\theta_{j'}\,.\\
\end{align*}
Hence, the induced metric on $S^{n-1}$  embedded in $\mathbb{R}^{n}$ is
$$
g_{\mu\nu}= \left(\lambda^T\lambda\right)_{\mu\nu}.
$$
After some algebraic manipulations, it can be shown that the metric is in fact  diagonal
and its  determinant  takes the form
\begin{equation}\label{eq:detg}
g=\sin^{2(n-2)}\theta_{1}\sin^{2(n-3)}\theta_{2}\cdots\sin^{4}\theta_{n-3}\sin^{2}\theta_{n-2}.
\end{equation}

The geodesic arc length $\theta$ between $\hat{x}$ and
$\hat{x}'$ on $S^{n-1}$ is  the angle given by
$$
\theta\equiv\arccos\hat{x}\cdot\hat{x}'=\arccos\sum_{i=1}^{n}\hat{x}_{i}\hat{x}_{i}'.
$$

\subsection{Laplacian in geodesic polar coordinates}
In geodesic polar coordinates $(r,\xi)$ around a point, one can show
using Equation~\ref{eq:Laplacian2} that the Laplacian on a
$d$-dimensional Riemannian manifold ${\mathcal M}$ takes the form
$$
\Delta = \partial_r^2 + (\partial_r \log \sqrt{g}) \partial_r + \Delta_{S_r^{d-1}},
$$
where $\Delta_{S_r^{d-1}}$ is the Laplacian induced on the geodesic
sphere $S_r^{d-1}$ of radius $r$.  If function $f$ depends only on the geodesic
distance $r$ from the fixed point, then
\begin{equation}
\Delta f(r)  =f''(r)+\left(\log\sqrt{g}\right)'f'(r),
\end{equation}
where $'$ denotes the radial derivative.

For the special case when ${\mathcal M}$ is  $S^{n-1}$, the coordinates
$\theta_1,\ldots,\theta_{n-1}$ described above correspond to the
geodesic polar coordinates around the north pole, with $r=\theta_1$. From
Equation~\ref{eq:detg}, we get
 
\begin{align*}
\log\sqrt{g(x)} & =(n-2)\log\sin r+(n-3)\log\sin\theta_{2}+\cdots\\
 & +\log\sin\theta_{n-2}.
\end{align*}
Note that only the first terms contributes to the radial derivative.
\subsection{Euclidean heat kernel}

Heat kernels in general are solutions to the heat equation
$$
\left(\partial_{t}-\Delta\right)\phi=0
$$
with a point-source (Dirac delta) initial condition. The heat kernel
in $\mathbb{R}^{d}$ is easily found to be
\begin{equation}\label{eq:euclideanHeatKernel}
G(\mathbf{x},\mathbf{y};t)=\left(\frac{1}{4\pi t}\right)^{\frac{d}{2}}K(\mathbf{x},\mathbf{y};t)
\end{equation}
where
$$
K(\mathbf{x},\mathbf{y};t)=\exp\left(-\frac{\|\mathbf{x}-\mathbf{y}\|^{2}}{4t}\right).
$$
$K$ is known as the Gaussian RBF kernel with parameter
$\gamma=1/4t$. $G(\mathbf{x},\mathbf{y};t)$ is the solution to the
heat equation satisfying the initial condition
$G(\mathbf{x},\mathbf{y};0)=\delta(\mathbf{x}-\mathbf{y})$.   Note
that formally,
$$
G(\mathbf{x},\mathbf{y};t) = {\rm e}^{t \Delta}  \delta(\mathbf{x}-\mathbf{y});
$$
using the Fourier transform representation of the right-hand side then
yields the expression in Equation~\ref{eq:euclideanHeatKernel}.

\subsection{Exact hyperspherical heat kernel}\label{appendix_proof}

We treat the hypersphere $S^{n-1}$ as being embedded in
$\mathbb{R}^{n}$ and use the induced metric on $S^{n-1}$ to define the
Laplacian.  The Laplacian in $\mathbb{R}^{n}$ takes the usual form

\begin{equation}\label{eq:LSeparation}
\Delta=\frac{1}{r^{n-1}}\partial_{r}\left(r^{n-1}\partial_{r}\right)-\frac{\hat{L}^{2}}{r^{2}}
\end{equation}
where the differential operator $\hat{L}^{2}$ depends only on the angular
coordinates. $-\hat{L}^{2}$  is the spherical Laplacian operator \cite{Wen:1985fm}.


\subsubsection{Spherical Laplacian and its eigenfunctions}
For $n=3$, the Laplacian  on  $\mathbb{R}^{3} $ is 
$$
\Delta=\frac{1}{r^{2}}\partial_{r}\left(r^{2}\partial_{r}\right)-\frac{\hat{L}^{2}}{r^{2}}
$$
where $\hat{L}^2$ is the squared orbital angular momentum operator in quantum
mechanics. Restricted to
$r=1$, the Laplacian reduces to the spherical Laplacian on $S^{2}$,
which is exactly the operator $-\hat{L}^{2}$ whose eigenfunctions
are the spherical harmonics $Y_{lm}(\theta,\phi)$ with eigenvalue
$-\ell(\ell+1)$. In this setting,  $Y_{lm}(\theta,\phi)$ can be viewed
as the angular component of homogeneous harmonic polynomials in $\mathbb{R}^3$, and
this perspective will be used in the subsequent discussion of
hyperspherical Laplacian. By convention, our spherical
harmonics satisfy the normalization condition
$$
\sum_{m=-\ell}^{\ell}|Y_{\ell m}(\theta,\phi)|^{2}=\frac{2\ell+1}{4\pi}
$$
and the completeness condition
$$
\sum_{\ell=0}^{\infty}\sum_{m=-\ell}^{\ell}Y_{\ell m}(\theta,\phi)Y_{\ell m}^{*}(\theta',\phi')  =  \delta(\cos\theta-\cos\theta')\delta(\phi-\phi').
$$

Analogous to the Euclidean case,  applying the evolution operator
$\exp(-\hat{L}^{2}t)$ on the initial delta distribution yields the
following eigenfunction expansion of the heat kernel on $S^{2}$:
$$
G(\hat{x},\hat{y};t)=\sum_{l=0}^{\infty}{\rm e}^{-\ell(\ell+1)t}\sum_{m=-\ell}^{\ell}Y_{\ell m}(\hat{x})Y_{\ell m}(\hat{y})^{*}.
$$
Applying the addition theorem of spherical harmonics,
$$
\frac{4\pi}{2\ell+1}\sum_{m=-\ell}^{\ell}Y_{\ell
  m}(\hat{x})Y_{\ell m}(\hat{y})^{*} = P_{\ell}(\hat{x}\cdot\hat{y}),
$$
we finally get
$$
G(\hat{x}\cdot\hat{y};t)=\sum_{\ell=0}^{\infty}\left(\frac{2\ell+1}{4\pi}\right){\rm e}^{-\ell(\ell+1)t}P_{\ell}(\hat{x}\cdot\hat{y}).
$$

\subsubsection{Generalization to $S^{n-1}$}
Similar to the spherical harmonics, the hyperspherical harmonics arise as the
angular part of degree-$\ell$ homogeneous harmonic polynomials
$h_{\ell}$ that satisfy $\Delta h_{\ell} =0$. In spherical coordinates
$(r,\xi)$, we can decompose $h_\ell (\mathbf{x})=r^{\ell}\tilde{Y}_{\ell}(\xi)$
\cite{Atkinson:2012tz,Wen:1985fm}, where $\tilde{Y}_{\ell}(\xi)$ is
the desired hyperspherical harmonic. Using the spherical coordinate
Laplacian in $\mathbb{R}^n$ shown in
Equation~\ref{eq:LSeparation}, we get
$$
0=\Delta h_{\ell}(\mathbf{x})=\tilde{Y}_{\ell}(\hat{x})\frac{1}{r^{n-1}}\partial_{r}\left(r^{n-1}\partial_{r}r^{\ell}\right)
- r^{\ell-2}\hat{L}^{2}\tilde{Y}_{\ell}(\xi),
$$
which can be simplified to yield the following eigenvalue equation for the
hyperspherical Laplacian:
$$
\hat{L}^{2}Y_{\ell\{m\}}=\ell(\ell+n-2)Y_{\ell\{m\}},
$$
where the set $\{m\}$ indexes the degenerate eigenstates.

\subsubsection{Lemmas for the proof of convergence}

To construct the eigenfunction expansion of the exact heat kernel and prove its convergence, we need
the following lemmas  \cite{Atkinson:2012tz,Wen:1985fm,Lorch:1984fw}:
\begin{lem}
The hyperspherical harmonics are complete on $S^{n-1}$ and resolve
the $\delta$-function
\begin{equation}
\delta(\hat{x},\hat{y})=\sum_{\ell=0}^{\infty}\sum_{\{m\}}Y_{\ell\{m\}}(\hat{x})Y_{\ell\{m\}}^{*}(\hat{y}).
\end{equation}
\end{lem}

\begin{lem}The hyperspherical harmonics  satisfy the
generalized addition theorem 
$$
\sum_{\{m\}}Y_{\ell\{m\}}(\hat{x})Y_{\ell\{m\}}(\hat{y})^{*}=\frac{1}{A_{S^{n-1}}}\frac{2\ell+n-2}{n-2}C_{\ell}^{\frac{n}{2}-1}(\hat{x}\cdot\hat{y}),
$$
where $C_{\ell}^{\nu}(w)$ are the Gegenbauer polynomials and $A_{S^{n-1}}=2\pi^{n/2}/\Gamma\left(\frac{n}{2}\right)$
is the surface area of $S^{n-1}$.
\end{lem}

\begin{lem}
\label{lem:The-Gegenbauer-polynomials}The Gegenbauer polynomials
$C_{\ell}^{\alpha}(w)$ with $\alpha>0$ and $\ell\ge0$ are bounded
in the interval $w\in[-1,1]$: in particular, $C_{0}^{\alpha}(w)=1$,
$C_{1}^{\alpha}(w)=\alpha w$, and thus, $|C_{1}^{\alpha}(w)|\le\alpha$
for $w\in[-1,1]$. Finally, for $\ell\ge2$, 
$$
|C_{\ell}^{\alpha}(w)|\le\left[w^{2}c_{2\ell,2\alpha}+(1-w^{2})c_{\ell,\alpha}\right],
$$
where 
$$
c_{\ell,\alpha}=\frac{\Gamma(\frac{\ell}{2}+\alpha)}{\Gamma(\alpha)\Gamma(\frac{\ell}{2}+1)}.
$$
\end{lem}

\subsubsection{The sweet spot of $t$}
Choosing an appropriate diffusion time $t$ for the heat kernel is important for machine
learning applications. Here, we use the degree of self-similarity
measured by the heat kernel as a function of $t$, and propose a choice
for which the self-similarity is neither too large nor too small.
If $t$ is too large, then the self-similarity is roughly  the
uniform similarity $1/A_{S^{n-1}}$, thereby losing contrast between neighbors
and outliers. By contrast, as $t$ approaches 0, the
self-similarity becomes infinite, and the sense of neighborhood
becomes too localized.  We thus need an intermediate value of $t$, for
which the self-similarity interpolates between the two limits.

The self-similarity is a special value of the heat kernel
\begin{eqnarray*}
G(1;t) & = & \sum_{\ell=0}^{\infty}{\rm e}^{-\ell(\ell+n-2)t}\frac{2\ell+n-2}{n-2}\frac{1}{A_{S^{n-1}}}C_{\ell}^{\frac{n}{2}-1}(1)\\
 & = & \frac{1}{A_{S^{n-1}}}\sum_{\ell=0}^{\infty}{\rm e}^{-\ell(\ell+n-2)t}\frac{2\ell+n-2}{n-2}\frac{\Gamma(\ell+n-2)}{\Gamma(\ell+1)\Gamma(n-2)}.
\end{eqnarray*}
Because the series converges rapidly for sufficiently large  $t$, we can truncate the series at
$\ell=\ell_{\max}$; i.e.
$$
G(1;t)
  \approx  \frac{1}{A_{S^{n-1}}}\sum_{\ell=0}^{\ell_{\max}}{\rm e}^{-\ell(\ell+n-2)t}\frac{2\ell+n-2}{n-2}\frac{\Gamma(\ell+n-2)}{\Gamma(\ell+1)\Gamma(n-2)}.
$$

\noindent In the large $n$ limit, we can bound the sum as
\begin{align*}
G(1;t) \leq \frac{1}{A_{S^{n-1}}}\sum_{\ell=0}^{\ell_{\max}}\left({\rm
    e}^{-nt}\right)^{\ell}\frac{n^{\ell}}{\ell!} \leq  \frac{\exp\left(n{\rm e}^{-nt}\right)}{A_{S^{n-1}}}.
\end{align*}
To keep the self-similarity finite, but larger than the uniform
similarity, suggests the choice for $t$ of order $\log n/n$, at which the self-similarity
is roughly  ${\rm e}/A_{S^{n-1}}$. We thus search for an optimal value of
$t$ around  $\log n/n$.

\subsubsection{SVM Classification\label{appendix_svm}}

In the main text, we denoted the parametrix and exact heat kernels
normalized by self-similarity as the ``parametrix kernel'' and ``exact
kernel,'' respectively. We then used the linear (lin), Gaussian RBF (rbf),
cosine (cos), parametrix (prx), and exact (ext) kernels in  SVM to (1)
classify WebKB-4-University web pages into four classes:
\emph{student}, \emph{faculty}, \emph{course}, and \emph{project}; and
(2) impute the binary mutation status of genes in TCGA-GBM data. The kernel
SVM classification results shown in the main text indicated that the
cosine kernel usually outperformed the linear kernel, most likely
as a pure consequence of the hyperspherical geometry, as we argue below.
The exact kernel outperformed the Gaussian RBF kernel for the WebKB
document data, but the advantage of exact kernel diminished in the
TCGA mutation count data. 
Figure 2 compares the
accuracy of SVM using different kernels on the TCGA-GBM data, where
the accuracy ratios rbf/lin, cos/lin, ext/lin, prx/cos, and ext/cos were
greater than 1 for most class sizes $m_{{\rm r}}$. Interestingly, the ratio
cos/lin showed some dependence on the sample size $m_{{\rm r}}$, and the exact kernel
also tended to outperform the Gaussian RBF kernel when $m_{{\rm r}}$
was small; in general, we noted that the hyperspherical kernels tended
to outperform the Euclidean kernels in small-sample-size
classification problems. This pattern may be understood by examining
the generalization error of kernel SVM as follows.

Intuitively, if a generic classifier were closely acquainted with the population
distribution of data through a large sample size, then its predictions
would be more generalizable to unseen samples. The ``largeness'' of
sample size $m$, however, is not explicitly quantifiable unless we
have a natural unit for it. Statistical learning theory
\cite{Vladimir:l-V4Juw7,Vapnik:1994dk,Guyon:1993ub} provides such a unit
associated with a probabilistic upper bound on generalization
errors.  That is, with probability at least $1-\eta$, the
generalization error of a binary SVM
classification is bounded from above by 
$$ F(\tilde{m};\mu_{{\rm VC}},\eta)=\sqrt{\frac{1}{\tilde{m}}\left[\left(\log2\tilde{m}+1\right)-\frac{\log\frac{\eta}{4}}{\mu_{{\rm VC}}}\right]} $$
where $\mu_{{\rm VC}}$ is the VC-dimension of the classifier, and
$\tilde{m}=m/\mu_{{\rm VC}}$ is the effective sample size.  The
derivative of $F(\tilde{m};\mu_{{\rm VC}},\eta)$ with respect to
$\tilde{m}$ is proportional to a positive factor times
$-\log\left[(2\tilde{m})^{\mu_{{\rm VC}}}4/\eta\right]$. Thus, the
upper bound decreases with $\tilde{m}$ when $(2\tilde{m})^{\mu_{{\rm
      VC}}}>\eta/4$, and increases otherwise; the critical effective
sample size $\tilde{m}_{{\rm
    crt}}=\frac{1}{2}\cdot(\eta/4)^{1/\mu_{{\rm
      VC}}}\approx\frac{1}{2}$ for typical values of $\mu_{{\rm VC}}>100$ and
$\eta\in[10^{-3},0.1]$. The VC dimension of a linear kernel SVM can be
estimated using an empirical upper bound
\cite{Vapnik:1994dk,Anonymous:2001vo}
$$
\mu_{VC}\le\mu_{VC}^{*}=\min\left\{ n,\frac{R^{2}}{M^{2}}\right\} +1,
$$
where $n$ is the feature space dimension, $R$ is the radius of the
smallest ball in feature space that encloses all data points, and $M$
is the SVM margin. We evaluated the bound $\mu_{{\rm VC}}^{*}$ for the
TCGA-GBM mutation count data with $C=1$, and found that the linear
kernel had $R^{2}/M^{2}\approx6\times10^{3}$ and thus that $\mu_{{\rm
    VC}}^{*{\rm lin}}=n+1\approx4\times10^{2}$. By contrast, the
cosine kernel, which is a linear kernel in the hyperspherically
transformed space with $R\le1$, had $\mu_{{\rm VC}}^{*{\rm cos}}$
approximately in the range $20\sim100\ll\mu_{{\rm VC}}^{*{\rm lin}}$,
as shown in Figure~3A. 
This reduction
in the VC-dimension is likely responsible for the classification
improvement of the cosine kernel over the linear kernel. We thus found
that $\tilde{m}_{\cos}=2m_{{\rm r}}/\mu_{{\rm VC}}^{*\cos}>\tilde{m}_{\rm crt}$, while
$\tilde{m}_{{\rm lin}}=2m_{{\rm r}}/\mu_{{\rm VC}}^{*{\rm lin}}< \tilde{m}_{\rm crt}$
for the TCGA-GBM data, and that the cosine kernel accuracy increased
with effective sample size, whereas the linear kernel accuracy tended
to decrease (Figure~3B,C, 
consistent with the analysis of
the upper bound on generalization error $F(\tilde{m};\mu_{{\rm
    VC}},\eta)$.  In addition, the Gaussian RBF and exact kernels
followed similar trends as the linear and cosine kernels, respectively
(Figure~3D,E
). Similar to the cosine kernel, the exact
kernel likely inherited the reduction in VC-dimension from the
hyperspherical map; as a result, the accuracy of the exact kernel also
increased with $\tilde{m}_{\cos}$, but with slightly higher accuracy
due to the additional tunable parameter $t$ that can adjust the
curvature of nonlinear decision boundaries. Moreover, the cases of
small sample size where the exact kernel outperformed the Gaussian RBF
kernel corresponded to the cases of larger effective sample size ratio
$\tilde{m}_{\rm cos}/ \tilde{m}_{\rm lin}$ (Figure~3
F).


\begin{thebibliography}{39}
\expandafter\ifx\csname natexlab\endcsname\relax\def\natexlab#1{#1}\fi
\expandafter\ifx\csname urlstyle\endcsname\relax
  \expandafter\ifx\csname doi\endcsname\relax
  \def\doi#1{doi:\discretionary{}{}{}#1}\fi \else
  \expandafter\ifx\csname doi\endcsname\relax
  \def\doi{doi:\discretionary{}{}{}\begingroup \urlstyle{rm}\Url}\fi \fi
\expandafter\ifx\csname selectlanguage\endcsname\relax
  \def\selectlanguage#1{}\fi

\bibitem{Lafferty:2005uy}
Lafferty J, Lebanon G.
\newblock {Diffusion Kernels on Statistical Manifolds}.
\newblock {\em Journal of Machine Learning Research\/} {\bf 6} (2005) 129--163.

\bibitem{Hastie:2013fd}
Hastie T, Tibshirani R, Friedman J.
\newblock {\selectlanguage{English}{\em {The Elements of Statistical
  Learning}\/}}.
\newblock Data Mining, Inference, and Prediction (Springer Science {\&}
  Business Media) (2013).
\newblock \doi{10.1111/j.1467-985X.2010.00646_6.x}.

\bibitem{Evgeniou:2001hc}
Evgeniou T, Pontil M.
\newblock {Support Vector Machines: Theory and Applications}.
\newblock {\em Machine Learning and Its Applications\/} (Berlin, Heidelberg:
  Springer Berlin Heidelberg) (2001), 249--257.
\newblock \doi{10.1007/3-540-44673-7_12}.

\bibitem{Boser:1992fz}
Boser BE, Guyon IM, Vapnik VN.
\newblock {\em {A training algorithm for optimal margin classifiers}\/} (New
  York, New York, USA: ACM) (1992).
\newblock \doi{10.1145/130385.130401}.

\bibitem{Cortes:1995fs}
Cortes C, Vapnik V.
\newblock {\selectlanguage{English}{Support-Vector Networks}}.
\newblock {\em Machine learning\/} {\bf 20} (1995) 273--297.
\newblock \doi{10.1023/A:1022627411411}.

\bibitem{Freund:1999dh}
Freund Y, Schapire RE.
\newblock {\selectlanguage{English}{Large Margin Classification Using the
  Perceptron Algorithm}}.
\newblock {\em Machine learning\/} {\bf 37} (1999) 277--296.
\newblock \doi{10.1023/A:1007662407062}.

\bibitem{Guyon:1993ub}
Guyon I, Boser B, Vapnik V.
\newblock {Automatic Capacity Tuning of Very Large VC-dimension Classifiers}.
\newblock {\em Advances in Neural Information Processing Systems\/}  (1993)
  147--155.

\bibitem{Kaufman:2009dk}
Kaufman L, Rousseeuw PJ.
\newblock {\selectlanguage{English}{\em {Finding Groups in Data}\/}}.
\newblock An Introduction to Cluster Analysis (Hoboken, NJ, USA: John Wiley
  {\&} Sons) (2009).
\newblock \doi{10.1002/9780470316801}.

\bibitem{Anonymous:U64WGXbS}
Belkin M, Niyogi P, Sindhwani V.
\newblock {Manifold Regularization: A Geometric Framework for Learning from
  Labeled and Unlabeled Examples}.
\newblock {\em Journal of Machine Learning Research\/} {\bf 7} (2006)
  2399--2434.

\bibitem{Aronszajn:1950bl}
Aronszajn N.
\newblock {Theory of reproducing kernels}.
\newblock {\em Transactions of the American mathematical society\/} {\bf 68}
  (1950) 337.
\newblock \doi{10.2307/1990404}.

\bibitem{Paulsen:2016wz}
Paulsen VI, Raghupathi M.
\newblock {\selectlanguage{English}{\em {An Introduction to the Theory of
  Reproducing Kernel Hilbert Spaces (Cambridge Studies in Advanced
  Mathematics)}\/}} (Cambridge University Press) (2016).

\bibitem{Berger:wXGLCovN}
Berger M, Gauduchon P, Mazet E.
\newblock {\em {Le spectre d'une variete riemannienne}\/} (Springer) (1971).

\bibitem{Hsu:2002vq}
Hsu EP.
\newblock {\em {Stochastic analysis on manifolds, volume 38 of Graduate Studies
  in Mathematics}\/} (American Mathematical Society) (2002).

\bibitem{Varopoulos:1987vx}
Varopoulos NT.
\newblock {\em {Random walks and Brownian motion on manifolds}\/} (Symposia
  Mathematica) (1987).

\bibitem{Ng:2002tj}
Ng A, Jordan M, Weiss Y, Dietterich T, Becker S.
\newblock {Advances in Neural Information Processing Systems, 14, chapter On
  spectral clustering: analysis and an algorithm} (2002).

\bibitem{Coifman:2006cy}
Coifman RR, Lafon S.
\newblock {\selectlanguage{English}{Diffusion maps}}.
\newblock {\em Applied and Computational Harmonic Analysis\/} {\bf 21} (2006)
  5--30.
\newblock \doi{10.1016/j.acha.2006.04.006}.

\bibitem{Atkinson:2012tz}
Atkinson K, Han W.
\newblock {\selectlanguage{English}{\em {Spherical Harmonics and Approximations
  on the Unit Sphere: An Introduction}\/}} (Springer Science {\&} Business
  Media) (2012).

\bibitem{Wen:1985fm}
Wen ZY, Avery J.
\newblock {\selectlanguage{English}{Some properties of hyperspherical
  harmonics}}.
\newblock {\em Journal of Mathematical Physics\/} {\bf 26} (1985) 396--9.
\newblock \doi{10.1063/1.526621}.

\bibitem{GOLDBART:2016uz}
Stone M, Goldbart P.
\newblock {Mathematics for physics: a guided tour for graduate students}.
\newblock Cambridge University Press, Cambridge (2009).
\newblock \doi{10.1017/CBO9780511627040}.

\bibitem{Grigoryan:1999du}
Grigor{\textquoteright}yan A.
\newblock {\selectlanguage{English}{Analytic and geometric background of
  recurrence and non-explosion of the Brownian motion on Riemannian
  manifolds}}.
\newblock {\em Bulletin of the American Mathematical Society\/} {\bf 36} (1999)
  135--249.
\newblock \doi{10.1090/S0273-0979-99-00776-4}.

\bibitem{Craven:1998tq}
Craven M, McCallum A, PiPasquo D, Mitchell T.
\newblock {Learning to extract symbolic knowledge from the World Wide Web}.
\newblock {\em Proceedings of the National Conference on Artificial
  Intelligence\/}  (1998) 509--516.

\bibitem{Hanahan:2011gu}
Hanahan D, Weinberg RA.
\newblock {\selectlanguage{English}{Hallmarks of Cancer: The Next Generation}}.
\newblock {\em Cell\/} {\bf 144} (2011) 646--674.
\newblock \doi{10.1016/j.cell.2011.02.013}.

\bibitem{Smedley:1999uq}
Smedley Dea.
\newblock {SHORT COMMUNICATION Cloning and Mapping of Members of the MYM
  Family}  (1999) 1--4.

\bibitem{Shchors:2002hu}
Shchors K, Yehiely F, Kular RK, Kotlo KU, Brewer G, Deiss LP.
\newblock {\selectlanguage{English}{Cell death inhibiting RNA (CDIR) derived
  from a 3'-untranslated region binds AUF1 and heat shock protein 27.}}
\newblock {\em Journal of Biological Chemistry\/} {\bf 277} (2002)
  47061--47072.
\newblock \doi{10.1074/jbc.M202272200}.

\bibitem{Zohrabian:2007wv}
Zohrabian VM, Nandu H, Gulati N.
\newblock {Gene expression profiling of metastatic brain cancer}.
\newblock {\em Oncology Reports\/}  (2007).

\bibitem{Kaur:2010jf}
Kaur B, Brat DJ, Calkins CC, Van~Meir EG.
\newblock {\selectlanguage{English}{Brain Angiogenesis Inhibitor 1 Is
  Differentially Expressed in Normal Brain and Glioblastoma Independently of
  p53 Expression}}.
\newblock {\em The American Journal of Pathology\/} {\bf 162} (2010) 19--27.
\newblock \doi{10.1016/S0002-9440(10)63794-7}.

\bibitem{Hamann:2015hv}
Hamann J, Aust G, Ara{\c c} D, Engel FB, Formstone C, Fredriksson R, et~al.
\newblock {\selectlanguage{English}{International Union of Basic and Clinical
  Pharmacology. XCIV. Adhesion G protein-coupled receptors.}}
\newblock {\em Pharmacological Reviews\/} {\bf 67} (2015) 338--367.
\newblock \doi{10.1124/pr.114.009647}.

\bibitem{Yamashita:2016dm}
Yamashita S, Fujii K, Zhao C, Takagi H, Katakura Y.
\newblock {\selectlanguage{English}{Involvement of the NFX1-repressor complex
  in PKC-$\delta$-induced repression of hTERT transcription}}.
\newblock {\em Journal of Biochemistry\/}  (2016) mvw038--5.
\newblock \doi{10.1093/jb/mvw038}.

\bibitem{Song:1994bh}
Song Z, Krishna S, Thanos D, Strominger JL, Ono SJ.
\newblock {\selectlanguage{English}{A novel cysteine-rich sequence-specific
  DNA-binding protein interacts with the conserved X-box motif of the human
  major histocompatibility complex class II genes via a repeated Cys-His domain
  and functions as a transcriptional repressor.}}
\newblock {\em Journal of Experimental Medicine\/} {\bf 180} (1994) 1763--1774.
\newblock \doi{10.1084/jem.180.5.1763}.

\bibitem{Adinolfi:2015kk}
Adinolfi E, Capece M, Franceschini A, Falzoni S.
\newblock {\selectlanguage{English}{Accelerated tumor progression in mice
  lacking the ATP receptor P2X7}}.
\newblock {\em Cancer research\/} {\bf 75} (2015) 635--644.
\newblock \doi{10.1158/0008-5472.CAN-14-1259}.

\bibitem{GomezVillafuertes:2015fd}
G{\'o}mez-Villafuertes R, Garc{\'\i}a-Huerta P, D{\'\i}az-Hern{\'a}ndez JI,
  Miras-Portugal MT.
\newblock {PI3K/Akt signaling pathway triggers P2X7 receptor expression as a
  pro-survival factor of neuroblastoma cells under limiting growth conditions}.
\newblock {\em Nature Publishing Group\/} {\bf 5} (2015) 1--15.
\newblock \doi{10.1038/srep18417}.

\bibitem{LinanRico:2016gu}
Li{\~n}{\'a}n-Rico A, Turco F, Ochoa-Cortes F, Harzman A, Needleman BJ,
  Arsenescu R, et~al.
\newblock {\selectlanguage{English}{Molecular Signaling and Dysfunction of the
  Human Reactive Enteric Glial Cell Phenotype}}.
\newblock {\em Inflammatory Bowel Diseases\/} {\bf 22} (2016) 1812--1834.
\newblock \doi{10.1097/MIB.0000000000000854}.

\bibitem{Anderton:2008gh}
Anderton JA, Lindsey JC.
\newblock {Global analysis of the medulloblastoma epigenome identifies
  disease-subgroup-specific inactivation of COL1A2}.
\newblock {\em Neuro-Oncology\/}  (2008).
\newblock \doi{10.1215/15228517-2008-048)}.

\bibitem{Liang:2007kw}
Liang Y, Diehn M, Bollen AW, Israel MA, Gupta N.
\newblock {\selectlanguage{English}{Type I collagen is overexpressed in
  medulloblastoma as a component of tumor microenvironment}}.
\newblock {\em Journal of Neuro-Oncology\/} {\bf 86} (2007) 133--141.
\newblock \doi{10.1007/s11060-007-9457-5}.

\bibitem{Schwalbe:2011ic}
Schwalbe EC, Lindsey JC, Straughton D, Hogg TL, Cole M, Megahed H, et~al.
\newblock {\selectlanguage{English}{Rapid diagnosis of medulloblastoma
  molecular subgroups.}}
\newblock {\em Clinical Cancer Research\/} {\bf 17} (2011) 1883--1894.
\newblock \doi{10.1158/1078-0432.CCR-10-2210}.

\bibitem{Vladimir:l-V4Juw7}
Vapnik VN.
\newblock {\selectlanguage{English}{\em {The Nature of Statistical Learning
  Theory}\/}} (Springer Science {\&} Business Media) (2013).

\bibitem{Vapnik:1994dk}
Vapnik V, Levin E, Le~Cun Y.
\newblock {\selectlanguage{English}{Measuring the VC-dimension of a learning
  machine}}.
\newblock {\em Neural Computation\/} {\bf 6} (1994) 851--876.
\newblock \doi{10.1162/neco.1994.6.5.851}.

\bibitem{Anonymous:2001vo}
Paliouras G, Karkaletsis V, Spyropoulos CD.
\newblock {\selectlanguage{English}{\em {Machine Learning and Its
  Applications}\/}}.
\newblock Advanced Lectures (Springer) (2003).

\bibitem{Lorch:1984fw}
Lorch L.
\newblock {\selectlanguage{English}{Inequalities for ultraspherical polynomials
  and the gamma function}}.
\newblock {\em Journal of Approximation Theory\/} {\bf 40} (1984) 115--120.
\newblock \doi{10.1016/0021-9045(84)90020-0}.

\end{thebibliography}
\end{document}